# Grounded Semantic Composition for Visual Scenes

**Peter Gorniak** PGORNIAK@MEDIA.MIT.EDU
**Deb Roy** DKROY@MEDIA.MIT.EDU
*MIT Media Laboratory*
*20 Ames St.,Cambridge, MA 02139 USA*

## Abstract

We present a visually-grounded language understanding model based on a study of how people verbally describe objects in scenes. The emphasis of the model is on the combination of individual word meanings to produce meanings for complex referring expressions. The model has been implemented, and it is able to understand a broad range of spatial referring expressions. We describe our implementation of word level visually-grounded semantics and their embedding in a compositional parsing framework. The implemented system selects the correct referents in response to natural language expressions for a large percentage of test cases. In an analysis of the system's successes and failures we reveal how visual context influences the semantics of utterances and propose future extensions to the model that take such context into account.

## 1. Introduction

We introduce a visually-grounded language understanding model based on a study of how people describe objects in visual scenes of the kind shown in Figure 1. We designed the study to elicit descriptions that would naturally occur in a joint reference setting and that are easy to produce and understand by a human listener. A typical referring expression for Figure 1 might be, "the far back purple cone that's behind a row of green ones". Speakers construct expressions to guide listeners' attention to intended objects. Such referring expressions succeed in communication because speakers and listeners find similar features of the visual scene to be salient, and share an understanding of how language is grounded in terms of these features. This work is a step towards our longer term goals to develop a conversational robot (Hsiao, Mavridis, & Roy, 2003) that can fluidly connect language to perception and action.

To study the use of descriptive spatial language in a task similar to one our robots perform, we collected several hundred referring expressions based on scenes similar to Figure 1. We analysed the descriptions by cataloguing the visual features that they referred to within a scene, and the range of linguistic devices (words or grammatical patterns) that they used to refer to those features. The combination of a visual feature and corresponding linguistic device is referred to as a *descriptive strategy.* The example sentence above contains several descriptive strategies that make use of colour, spatial relations, and spatial grouping. These descriptive strategies are used in composition by the speaker to make reference to a unique object.

We propose a set of computational mechanisms that correspond to the most commonly used descriptive strategies from our study. The resulting model has been implemented as a set of visual feature extraction algorithms, a lexicon that is grounded in terms of these visual

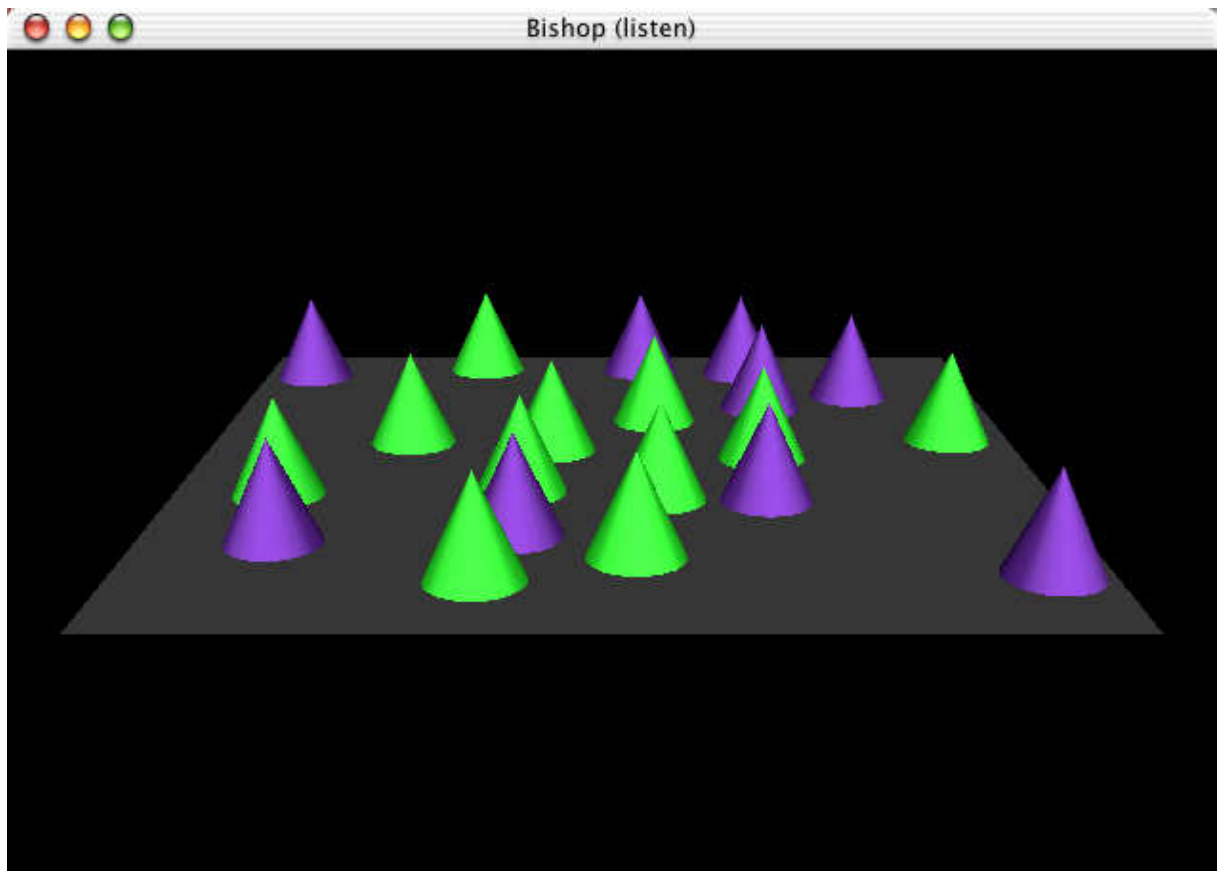

Figure 1: A sample scene used to elicit visually-grounded referring expressions (if this figure has been reproduced in black and white, the light cones are green in colour, the dark cones are purple)

features, a robust parser to capture the syntax of spoken utterances, and a compositional engine driven by the parser that combines visual groundings of lexical units. To evaluate the system, we collected a set of spoken utterances from three speakers. The verbatim transcriptions of the speech, complete with speech repairs and various other ungrammaticalities common in spoken language, were fed into the model. The model was able to correctly understand the visual referents of 59% of the expressions (chance performance assuming that a random object is selected on each of a session's 30 trials was $1/30 \sum_{i=1}^{30} 1/i = 13\%$). The system was able to resolve a range of linguistic phenomena that made use of relatively complex compositions of spatial semantics. We provide a detailed analysis of the sources of failure in this evaluation, based on which we propose a number of improvements that are required to achieve human level performance. In designing our framework we build on prior work in human reference resolution and integration of semantics during parsing. The main contribution of this work lies in using visual features based on a study of human visual and linguistic reference as the grounded semantic core of a natural language understanding system.

While our previous work on visually-grounded language has centred on machine learning approaches (Roy, Gorniak, Mukherjee, & Juster, 2002; Roy & Pentland, 2002; Roy, 2002), we chose not to apply machine learning to the problem of compositional grounded semantics in this investigation. Rather, we endeavoured to provide a framework that can process the types of descriptive strategies and compositionality that were found in our study with human participants. In future work, we will investigate how machine learning methods can be used to acquire parts of this framework from experience, leading to more robust and accurate performance.

### 1.1 Grounded Semantic Composition

We use the term *grounded semantic composition* to highlight that both the semantics of individual words and the word composition process itself are visually-grounded. In our model, each lexical entry's meaning is grounded through an association to a visual model. For example, "green" is associated with a probability distribution function defined over a colour space. We propose processes that combine the visual models of words, governed by rules of syntax.

Given our goal of understanding and modelling grounded semantic composition, several questions arise:

- What are the visual features that people use to describe objects in scenes such as Figure 1?
- How do these features connect to language?
- How do features and their descriptions combine to produce whole utterances and meanings?
- Are word meanings independent of the visual scene they describe?
- Is the meaning of the whole utterance based only on the meanings of its parts?
- Is composition of meanings a purely incremental process?

We assumed "easy" answers to some of these questions as a place to start our modelling effort. Our current implementation assumes that the meaning of a whole utterance is fully derived from the meanings of its parts, performs composition incrementally (without backtracking), and does not let the visual context influence the interpretation of word meanings. Despite these assumptions, the system handles relatively sophisticated semantic composition. When evaluated on test data, the system correctly understood and chose appropriate referents for expressions such as, "the purple one behind the two green ones" and "the left green cone in front of the back purple one".

After analysing our system's performance on human participants' utterances, we found that:

- Word meanings can be strongly dependent on the visual scene they describe. For instance, we found four distinct visual interpretations for the word "middle" that are linguistically indistinguishable, but instead depend on different visual contexts to be understood.
- The meaning of an utterance may sometimes depend on more than the meanings of its parts. Its meaning may also depend on the visual context in which the utterance occurs, which can modify how parts compose. For example, some objects that are referred to as "frontmost left" would be referred to neither as "left" or "frontmost" in isolation, nor are they the result of a multiplicative joined estimation of the two.
- Composition of meanings in this task is not a purely incremental process. In some cases we found it necessary to backtrack and reinterpret parts of the utterance when no good referents can be found at a later processing stage, or when ambiguities cannot be resolved with the current interpretation. Due to a strictly feed forward model of language understanding, our current implementation fails on such cases.

These results are similar to those reported in prior studies (Brown-Schmidt, Campana, & Tanenhaus, 2002; Griffin & Bock, 2000; Pechmann, 1989). Although our model does not currently address these issues of context-dependent interpretation and backtracking, we believe that the framework and its approach to grounded compositional semantics provide useful steps towards understanding spatial language. The system performs well in understanding spatial descriptions, and can be applied to various tasks in natural language and speech based human-machine interfaces.

This paper begins by highlighting several strands of related previous work. In section 2, we introduce the visual description task that serves as the basis for this study and model. Section 3 presents our framework for grounded compositional semantics. Section 4 describes the resulting computational model. An example of the whole system at work is given in Section 5. We discuss the results of applying our system to human data from the spatial description task in section 6, together with an analysis of the system's successes and failures. This leads to suggestions for future work in Section 7, followed by a summary in section 8.

### 1.2 Related Work

Winograd's SHRDLU is a well known system that could understand and generate natural language referring to objects and actions in a simple blocks world (Winograd, 1970). Like our system it performs semantic interpretation during parsing by attaching short procedures to lexical units (see also Miller & Johnson-Laird, 1976). However, SHRDLU had access to a clean symbolic representation of the scene, whereas the system discussed here works with a synthetic vision system and reasons over geometric and other visual measures. Furthermore, we intend our system to robustly understand the many ways in which human participants verbally describe objects in complex visual scenes to each other, whereas SHRDLU was restricted to sentences it could parse completely and translate correctly into its formalism.

SHRDLU is based on a formal approach to semantics in which the problem of meaning is addressed through logical and set theoretic formalisms. Partee provides an overview of this approach and to the problems of context based meanings and meaning compositionality from this perspective (Partee, 1995). Our work reflects many of the ideas from this work, such as viewing adjectives as functions. Pustejovsky's theory of the Generative Lexicon (GL) in particular takes seriously noun phrase semantics and semantic compositionality (Pustejovsky, 1995). Our approach to lexical semantic composition was originally inspired by Pustejovsky's qualia structures. However, these formal approaches operate in a symbolic domain and leave the details of non-linguistic influences on meaning unspecified, whereas we take the computational modelling of these influences as our primary concern.

Research concerning human production of referring expressions has lead to studies related to the one described here, but without computational counterparts. Brown-Schmidt e.a., for example, engage participants in a free-form dialogue (as opposed to the one-sided descriptions in our task) producing referential descriptions to solve a spatial arrangement problem (Brown-Schmidt et al., 2002). Due to the use of more complicated scenes and complete dialogues, they find that their participants often engage in agreement behaviours and use discourse and visual context to disambiguate underspecified referring expressions more often than in our study. Similar tasks have been used in other studies of dialogue and referring expressions (Pechmann, 1989; Griffin & Bock, 2000). We intentionally eliminated dialogue and used a simpler visual scene and task to elicit spatial descriptions (as opposed to description by object attributes), and to be able to computationally model the strategies our participants employ. Formal theories of vagueness support our findings in the expressions produced by our participants (Kyburg & Morreau, 2000; Barker, 2002).

Word meanings have been approached by several researchers as a problem of associating visual representations, often with complex internal structure, to word forms. Models have been suggested for visual representations underlying colour (Lammens, 1994) and spatial relations (Regier, 1996; Regier & Carlson, 2001). Models for verbs include grounding their semantics in the perception of actions (Siskind, 2001), and grounding in terms of motor control programs (Bailey, 1997; Narayanan, 1997). Object shape is clearly important when connecting language to the world, but remains a challenging problem in computational models of language grounding. In previous work, we have used histograms of local geometric features which we found sufficient for grounding names of basic objects (dogs, shoes, cars, etc.) (Roy & Pentland, 2002). This representation captures characteristics of the overall outline form of an object that is invariant to in-plane rotations and changes of scale. Landau and Jackendoff provide a detailed analysis of additional visual shape features that play a role in language (Landau & Jackendoff, 1993). For example, they suggest the importance of extracting the geometric axes of objects in order to ground words such as "end", as in "end of the stick". Shi and Malik propose an approach to performing visual grouping on images (Shi & Malik, 2000). Their work draws from findings of Gestalt psychology that provide many insights into visual grouping behaviour (Wertheimer, 1999; Desolneux, Moisan, & Morel, 2003). Engbers e.a. give an overview and formalization of the grouping problem in general and various approaches to its solution (Engbers & Smeulders, 2003). In parallel with the work presented in this paper, we also been studying visual grouping and will fold the results into the systen described here (Dhande, 2003).

Our model of incremental semantic interpretation during parsing follows a tradition of employing constraint satisfaction algorithms to incorporate semantic information starting with SHRDLU and continued in other systems (Haddock, 1989). Most prior systems use a declaratively stated set of semantic facts that is disconnected from perception. Closely related to our work in this area is Schuler's (2003), who integrates determination of referents to the parsing process by augmenting a grammar with logical expressions, much like we augment a grammar with grounded composition rules (see Section 3.4). Our emphasis, however, is on a system that can actively ground word and utterance meanings through its own sensory system. Even though the system described here senses a synthetic scene, it makes continuous measurements during the parsing process and we are now integrating it into an active vision system (Hsiao et al., 2003). Schuler's system requires a human-specified clean logical encoding of the world state, which ignores the noisy, complex and difficult-to-maintain process linking language to a sensed world. We consider this process, which we call the grounding process, one of the most important aspects of situated human-like language understanding.

SAM (Brown, Buntschuh, & Wilpon, 1992) and Ubiquitous Talker (Nagao & Rekimoto, 1995) are language understanding systems that map language to objects in visual scenes.

Similar to SHDRLU, the underlying representation of visual scenes is symbolic and loses much of the subtle visual information that our work, and the work cited above, focus on. Both SAM and Ubiquitous Talker incorporate a vision system, phrase parser and understanding system. The systems translate visually perceived objects into a symbolic knowledge base and map utterances into plans that operate on the knowledge base. In contrast, we are primarily concerned with understanding language referring to the objects and their relations as they appear visually.

We have previously proposed methods for visually-grounded language learning (Roy & Pentland, 2002), understanding (Roy et al., 2002), and generation (Roy, 2002). However, the treatment of semantic composition in these efforts was relatively primitive. For a phrase, the visual models of each word in the phrase were individually evaluated and multiplied. This method only works for phrases with conjunctive modifiers, and even in those cases, as we discuss later, ordering of modifiers sometimes needs to be taken into account (i.e., "leftmost front" does not always refer to what "front leftmost" does). While this simple approach worked in the constrained domains that we have addressed in the past, it does not scale to the present task. For example, the Describer system (Roy, 2002) encodes spatial locations in absolute terms within the frame of reference of a visual scene. As a result, Describer makes mistakes that humans would not make. Its grounding of the word "highest", as an example, is defined by a probability distribution centred at a specific height in the scene, so that the object closest to that height is the best example of "highest", not accounting for the fact that there may be objects at greater height (depending on the relative sizes and shapes of objects). In addition, Describer's only interpretation of a phrase like "the highest green rectangle" is to find an object that is both close to the center of the probability distributions for "highest" and for "green", not accounting for the fact that for a human listener the highest green rectangle need not be high on the screen at all (but only higher than the other green rectangles). A word such as "highest" requires a visual binding that includes some notion of rank ordering. Such a move, however, requires a rethinking of how to achieve semantic composition, which is addressed in the approach here.

## 2. A Spatial Description Task

We designed a task that requires people to describe objects in computer generated scenes containing up to 30 objects with random positions on a virtual surface. The objects all had identical shapes and sizes, and were either green or purple in colour. Each object had a 50% chance of being green, otherwise it was purple. We refer to this task as the Bishop task, and to the resulting language understanding model and implemented system simply as Bishop.

### 2.1 Motivation for Task Design

In our previous work, we have investigated how speakers describe objects with distinctive attributes like colour, shape and size in a constraint speaking task and in scenes with a constant number of objects (Roy, 2002). Speakers in such a task are rarely compelled to use spatial relations and never use groups of objects, because in most cases objects can be distinguished by listing their properties. In designing the Bishop task, our goal was to naturally lead speakers to make reference to spatial aspects of the scene. Therefore, we drastically increased the number of objects in the scene and decreased the number of distinctive object attributes. We also let the number of objects vary throughout the trials to cover both scenes cluttered with objects and scenes with only a few objects in our analysis.

In a variation of the task, we ran experiments in which the system chose objects at random for the speaker to describer, rather than allowing the describer to make the choice. We found that this made the task difficult and highly unnatural for the speaker as there were often few visually salient arrangements that the randomly chosen objects took part in. As a result, listeners make many more errors in resolving the reference in this variation of the task (3.5% error when the speaker chose the object versus 13% when the system chose). There are limits to the accuracy of pure linguistic reference which we appeared to be reaching in the random selection version of the task. Speakers seemed to have a much harder time finding visually salient landmarks, leading to long and less natural descriptions, for example in the centre there are a bunch of green cones, four of them, um, actually there are more than four, but, ah, theres one thats in the centre pretty much of the pile of them up to the its at the top, ahm, how can you say this... or the seventh cone from the right side (followed by the listener counting cones by pointing at the screen). To avoid collecting such unnatural data, we decided not to use the random selection version of the task.

Another possible variant of the task would be to let the system choose objects in some non-random manner based on the system's analysis of which objects would be more natural to describe. However, this approach would clearly bias the data towards objects that matched preexisting models of the system.

Since we are interested in how people described objects spatially as well as which visual features they found salient, we decided to let the listener pick objects that he or she felt were concisely yet not trivially describable. We acknowledge that this task design eases the difficulty of the understanding task; when speakers could not find an interesting object that was easy to describe in other ways, they resorted to simpler choices like "the leftmost one". Yet, the utterances elicited through this task are relatively complex (see Appendix A for a complete listing) and and provided serious challenges from an automatic language understanding perspective.

Scenes were rendered in 3D instead of using an equivalent 2D scene in anticipation of the transition of the understanding system to a camera driven vision system. The use of 3D rendering introduces occlusion, shadows, and other sources of ambiguity that must eventually be addressed if we are to transition to a real vision system. However, we note that the scene did not include any interesting 3D spatial relations or other features, and that we do not claim that the description task and thus the system presented here would generalize directly to a true 3D setting. Furthermore, we did not use any 3D information about the visual scene, so our system interprets all spatial relations in 2D. This errs on the 2D side of the ambiguity inherent in a word like "leftmost" in reference to one of our scenes (the interpretation can differ due to perspective effects: the "leftmost" object when interpreting the scene as 2D might not be the "leftmost" when interpreting it as 3D. We believe the task and the types of visually grounded descriptions it produced were challenging for a computational system to understand, as we hope to show in the remainder of this paper.

Finally, we should note that our goal was not to design a task to study collaboration, dialogue and agreement, which is the goal of other experiments and analyses (Carletta & Mellish, 1996; Eugenio, Jordan, Thomason, & Moore, 2000). We use a speaker/listener dyad to ensure that the descriptions produced are understandable to a human listener, but we purposefully did not allow listeners to speak. Their only feedback channel to speakers was the successful or unsuccessful selection of the described object. While this does introduce a minimal form of dialogue, the low error rate of listeners leads us to believe that negative reinforcement was negligible for all speakers and that the task should not be viewed as an exercise in collaboration. We cannot rule out that listeners adopted strategies used by their partners when it was their turn to speak. However, the relative similarity of strategies between pairs shows that this phenomenon does not make the data unrepresentative, and even produces the types of shortenings and vagueness that we would expect to see in an extended description task when speaking to a machine.

### 2.2 Data Collection

Participants in the study ranged in age from 22 to 30 years, and included both native and non-native English speakers. Pairs of participants were seated with their backs to each other, each looking at a computer screen which displayed identical scenes such as that in Figure 1. In each pair, one participant served as describer, and the other as listener. The describer wore a microphone that was used to record his or her speech. The describer used a mouse to select an object from the scene, and then verbally described the selected object to the listener. The listener was not allowed to communicate verbally or otherwise at all, except through object selections. The listener's task was to select the same object on their own computer display based on the verbal description. If the selected objects matched, they disappeared from the scene and the describer would select and describe another object. If they did not match, the describer would re-attempt the description until understood by the listener. Using a describer-listener dyad ensured that speech data resembled natural communicative dialogue. Participants were told they were free to select any object in the scene and describe it in any way they thought would be clear. They were also told not to make the task trivial by, for example, always selecting the leftmost object and describing it as "leftmost". The scene contained 30 objects at the beginning of each session, and a session ended when no objects remained, at which point the describer and listener switched roles and completed a second session (some participants fulfilled a role multiple times). We found that listeners in the study made extremely few mistakes in interpreting descriptions, and seemed to generally find the task easy to perform.

Initially, we collected 268 spoken object descriptions from 6 participants. The raw audio was segmented using our speech segmentation algorithm based on pause structure (Yoshida, 2002). Along with the utterances, the corresponding scene layout and target object identity were recorded together with the times at which objects were selected. This 268 utterance corpus is referred to as the development data set. We manually transcribed each spoken utterance verbatim, retaining all speech errors (false starts and various other ungrammaticalities). Rather than working with grammatically controlled language, our interest was to model language as it occurs in conversational settings since our longer term goal is to "transplant" the results of this work into conversational robots where language will be in spoken form. Off-topic speech events (laughter, questions about the task, other remarks, and filled pauses) were marked as such (they do not appear in any results we report).

We developed a simple algorithm to pair utterances and selections based on their time stamps. This algorithm works backwards in time from the point at which the correct object was removed from the scene. It collects all on-topic utterances that occurred before this removal event and after the previous removal event and that are not more than 4 seconds apart. It fuses them into a single utterance, and sends the scene description, the complete utterance and the identity of the removed object to the understanding system. The utterance fusing is necessary because participants often paused in their descriptions. At the same time, pauses beyond a certain length usually indicated that the utterances before the pause contained errors or that a rephrase occurred. This pairing algorithm is obviously of a heuristic nature, and we mark instances where it makes mistakes (wrongly leaving out utterances or attributing utterances to the wrong selection event) in the analysis of our data below. When we report numbers of utterances in data sets in this paper, they correspond to how many utterance-selection pairs our pairing algorithm produces. This means that due to errors by this algorithm the numbers of utterances we report are not divisible by 30, the actual number of objects selected in each session.

The development corpus was analysed to catalogue the range of common referring strategies (see Section 2.3). This analysis served as a basis for developing a visually-grounded language understanding system designed to replace the human listener in the task described above. Once this implementation yielded acceptable results on the development corpus, we collected another 179 spoken descriptions from three additional participants to evaluate generalization and coverage of our approach. We used exactly the same equipment, instructions and collection protocol as in collecting the development data to collect the test data. The average length of utterances in both the development and the test set was between 8 and 9 words. The discussion and analysis in the following sections focuses on the development set. In Section 6 we discuss performance on the test set.

### 2.3 Descriptive Strategies for Achieving Joint Reference

As we noted earlier, we call a combination of a visual feature measured on the current scene (or, in the case of anaphora, on the previous scene) together with its linguistic realization a descriptive strategy. In this section, we catalogue the most common strategies that describers used to communicate to listeners. This analysis is based strictly on the development data set. We discuss how our implemented system handles these categories in Section 4.

We distinguish three subsets of our development data:

- A set containing those utterance/selection pairs that contain errors. An error can be due to a repair or mistake on the human speaker's part, a segmentation mistake by our speech segmenter, or an error by our utterance/selection pairing algorithm.

- A set that contains those utterance/selection pairs that employ descriptive strategies other than those we cover in our computational understanding system (we cover those in Sections 2.3.1 to 2.3.5).

- The set of utterance/selection pairs in the development data that are not a member of either subset described above. We refer to this subset as the 'clean' set.

Note that the first two subsets are not mutually exclusive. In the following sections, we report two percentages for each descriptive strategy. The first is the percentage of utterance/selection pairs that employ a specific descriptive strategy relative to all the utterance/selection pairs in the development data set. The second is the percentage of utterance/selection pairs relative to the clean set of utterance/selection pairs, as described above. All examples given in this paper are actual utterances and scenes from our development and test sets.

### 2.3.1 Colour

Almost every utterance employs colour to pick out objects. While designing the task, we intentionally trivialized the problem of colour reference. Objects come in only two distinct colours, green and purple. Unsurprisingly, all participants used the terms "green" and "purple" to refer to these colours. In previous work we have addressed the problems of learning visually-grounded models of colour words (Roy & Pentland, 2002; Roy, 2002). Here, our focus is semantic compositionality of terms, so we chose to simplify the colour naming problem. Figure 2 shows one of the few instances where only colour is used to pick out a referent. Most of the examples in subsequent sections will be of colour composed with other descriptive strategies.

Syntactically, colours are expressed through adjectives (as mentioned: "green" and "purple") that always directly precede the nouns they modify. That is, nobody ever said "the green left one" in the data, but rather adjectives would commonly occur in the order "the left green one".

In our data, "green" and "purple" can also sometimes take the roles of nouns, or at least be left dangling in a noun phrase with an ellipse like "the leftmost purple". Although this form of dangling modifier might seem unlikely, it does occur in spoken utterances in our task. As the only objects in Bishop are cones, participants had no trouble understanding such ellipsis, which occur in 7% of the data.

Participants used colour to identify one or more objects in 96% of the data, and 95% of the clean data.

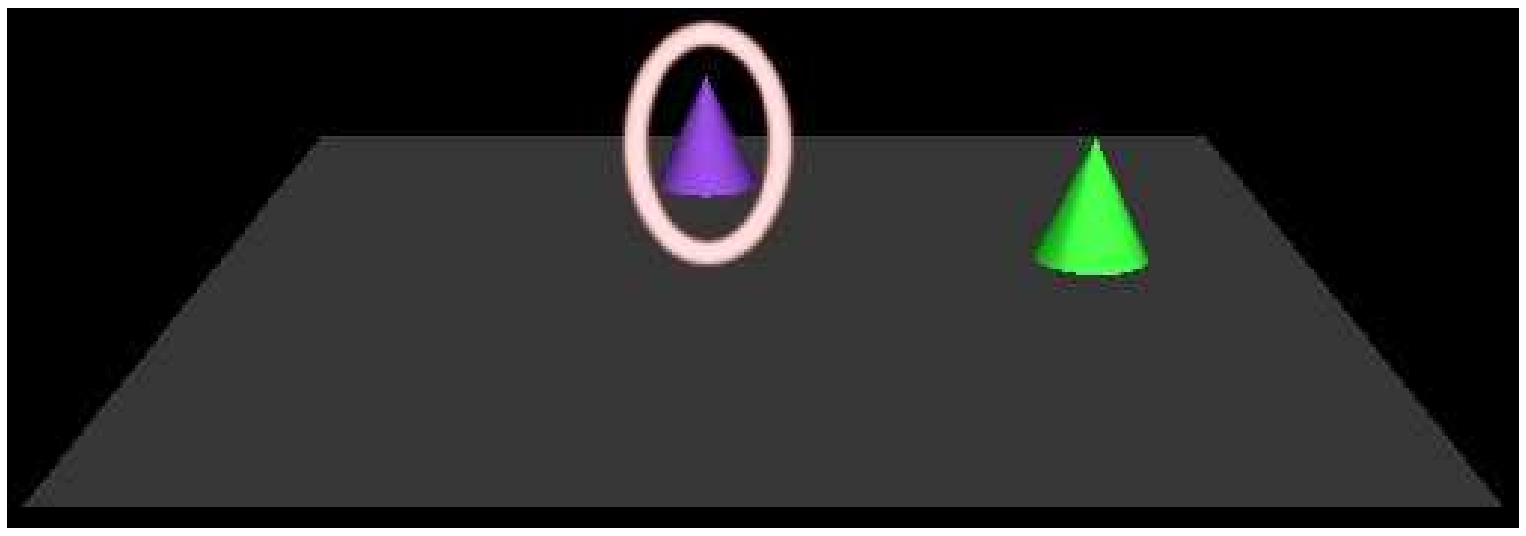

"the purple cone"

Figure 2: Example utterance using only colour

### 2.3.2 Spatial Regions and Extrema

The second most common descriptive strategy is to refer to spatial extremes within groups of objects and to spatial regions in the scene. The left example in Figure 3 uses two spatial terms to pick out its referent: “closest” and “in the front”, both of which leverage spatial extrema to direct the listener’s attention. In this example, selection of the spatial extremum appears to operate relative to only the green objects, i.e. the speaker seems to first attend to the set of green cones, then choose amongst them. Alternatively, “closest” and “in the front” could pick several objects of any colour, and the colour specification could then filter these spatial extrema to determine a final referent. In this case the two interpretations yield the same referent, but there are cases in the corpus in which the second alternative (spatial selection followed by colour selection) yields no referents at all.

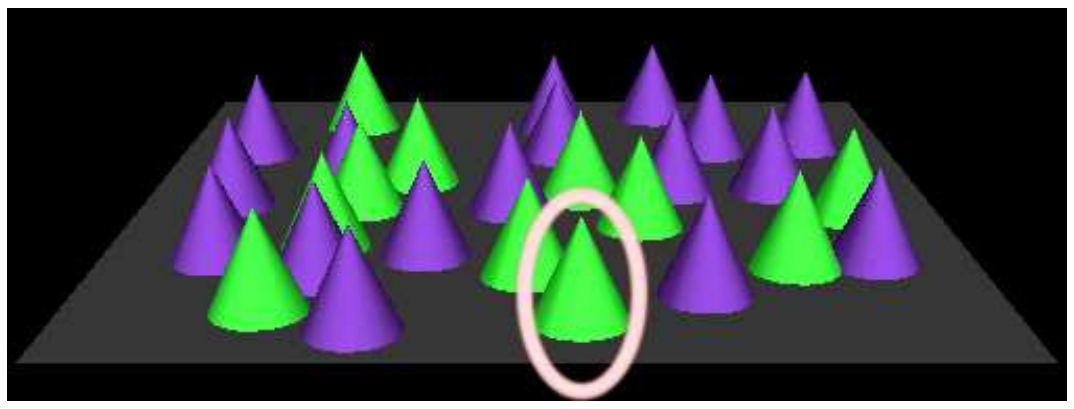

“the green one that’s closest to us in the front”

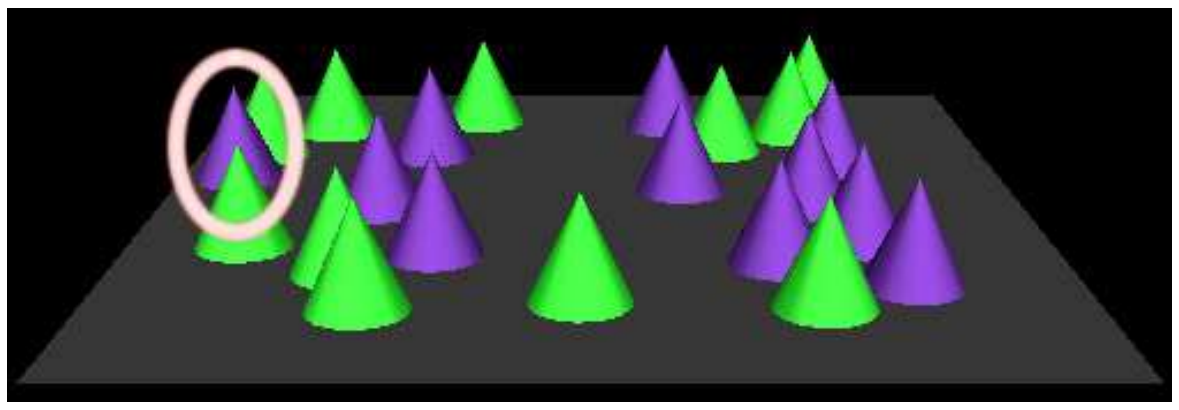

“the purple one on the left side”

Figure 3: Example utterances specifying objects by referring to spatial extrema

The right example in Figure 3 shows that phrases not explicitly indicating spatial extrema are still sometimes intended to be interpreted as referring to extrema. If the listener was to interpret “on the left side” as referring to the left side of the scene, the phrase would be ambiguous since there are four purple cones on the left side of the scene. On the other hand, the phrase is unambiguous if interpreted as picking out an extremum. Figure 4 shows an instance where “on the right hand side” actually refers to a region on the board. The first example in that figure shows the phrase “on the right hand side” combined with an extremum term, “lowest”. Note that the referent is not the right extremum. In the second example in Figure 4, the referent is not the “bottommost” green object, and, (arguably, if taking the scene as existing in 3D), neither is it the “leftmost”. Regions on the board seem to play a role in both cases. Often the local context of the region may play a stronger role than the global one, as the referent in the second example in Figure 4 can be found by attending to the front left area of the scene, then selecting the “left bottom” example amongst the candidates in this area. Along the same lines, words like “middle” are largely used to describe a region on the board, not a position relative to other cones.

Being rather ubiquitous in the data, spatial extrema and spatial regions are often used in combination with other descriptive strategies like grouping, but are most frequently combined with other extrema and region specifications. As opposed to when combined with colour adjectives, multiple spatial specifications tend to be interpreted in left to right order, that is, selecting a group of objects matching the first term, then amongst those choosing objects that match the second term. The examples in Figure 4 could be understood as

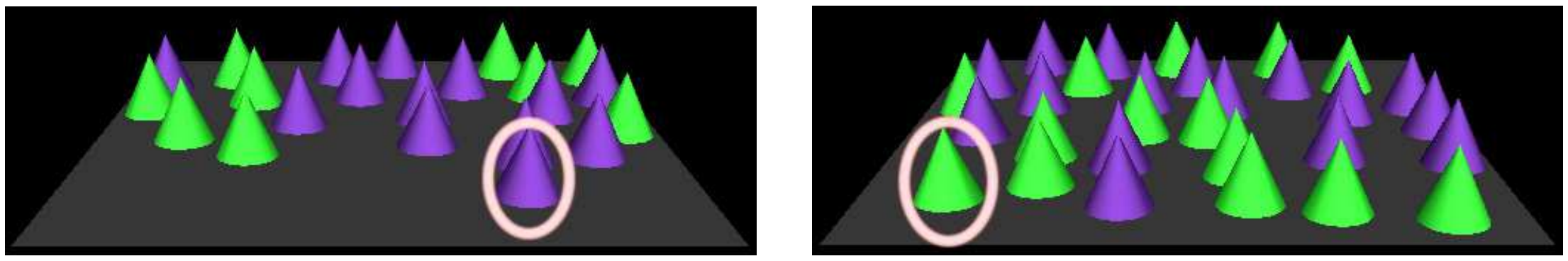

"the lowest purple on the right hand side" "the green cone at the left bottom"

Figure 4: Example utterances specifying regions

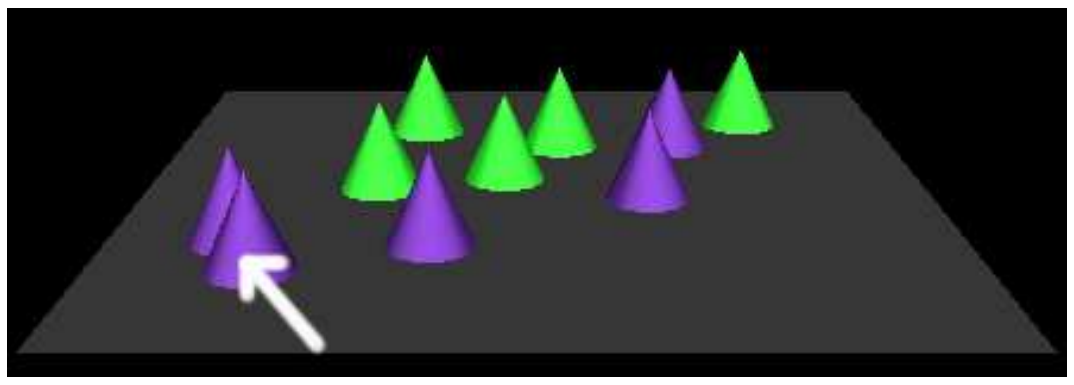

"the purple one in the front left corner"

Figure 5: Extrema in sequence

simply ignoring the order of spatial specifications and instead finding a conjoined best fit, i.e. the best example of both "bottommost" and "leftmost". However, Figure 5 demonstrates that this is not generally the case. This scene contains two objects that are best fits to an unordered interpretation of "front left", yet the human participant confidently picks the front object. Possible conclusions are that extrema need to be be interpreted in sequence, or that participants are demonstrating a bias preferring front-back features over left-right ones. In our implementation, we choose to sequence spatial extrema in the order they occur in the input.

Participants used single spatial extrema to identify one or more objects in 72% of the data, and in 78% of the clean data. They used spatial region specifications in 20% of the data (also 20% of the clean data), and combined multiple extrema or regions in 28% (30% of the clean data).

#### 2.3.3 Grouping

To provide landmarks for spatial relations and to specify sets of objects to select from, participants used language to describe groups of objects. Figure 6 shows two examples of such grouping constructs, the first using an unnumbered group of cones ("the green cones"), the second using a count to specify the group ("three"). The function of the group is different in the two examples: in the left scene the participant specifies the group as a landmark to serve in a spatial relation (see Section 2.3.4), whereas in the right scene the participant first specifies a group containing the target object, then utters another description to select within that group. Note that grouping alone never yields an individual reference, so participants compose grouping constructs with further referential tactics (predominantly extrema and spatial relations) in all cases.

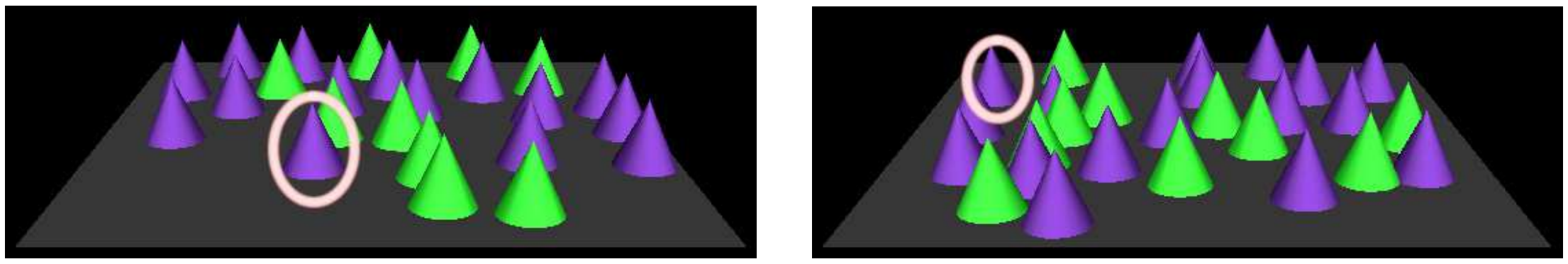

"the purple cone in the middle to the left of the green cones"

"there's three on the left side; the one in the furthest back"

Figure 6: Example utterances using grouping

Participants used grouping to identify objects in 12% of the data and 10% of the clean data. They selected objects within described groups in 7.5% of the data (8% of the clean data) and specified groups by number of objects ("two", "three", ...) in 8.5% of the data (also 8.5% of the clean data).

#### 2.3.4 Spatial Relations

As already mentioned in Section 2.3.3, participants sometimes used spatial relations between objects or groups of objects. Examples of such relations are expressed through prepositions like "below" or "behind" as well as phrases like "to the left of" or "in front of". We already saw an example of a spatial relation involving a group of objects in Figure 6, and Figure 7 further shows two examples that involve spatial relations between individual objects. The first example is one of the few examples of 'pure' spatial relations between two individual objects referenced only by colour. The second example is a more typical one where the spatial relation is combined with another strategy, here an extremum (as well as two speech errors by the describer).

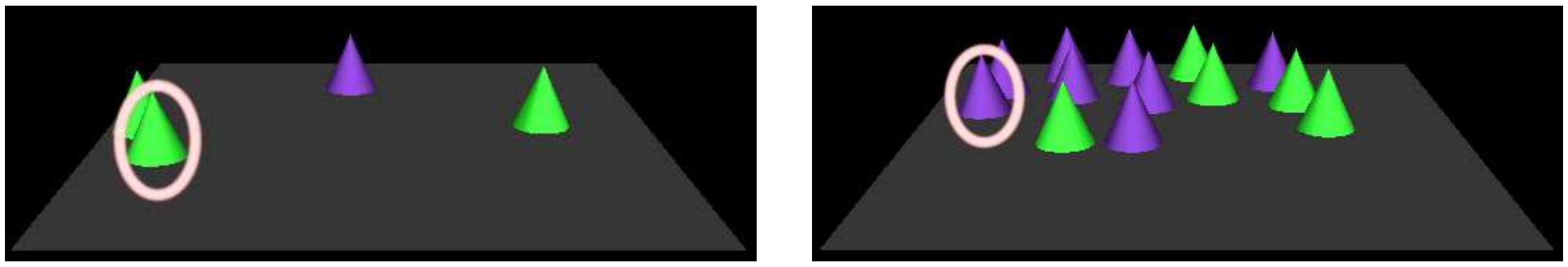

"the green cone below the green cone"

"there's a purple cone that's it's all the way on the left hand side but it's it's below another purple"

Figure 7: Example utterances specifying spatial relations

Participants used spatial relations in 6% of the data (7% of the clean data).

### 2.3.5 Anaphora

In a number of cases participants used anaphoric references to the previous object removed during the description task. Figure 8 shows a sequence of two scenes and corresponding utterances in which the second utterance refers back to the object selected in the first.

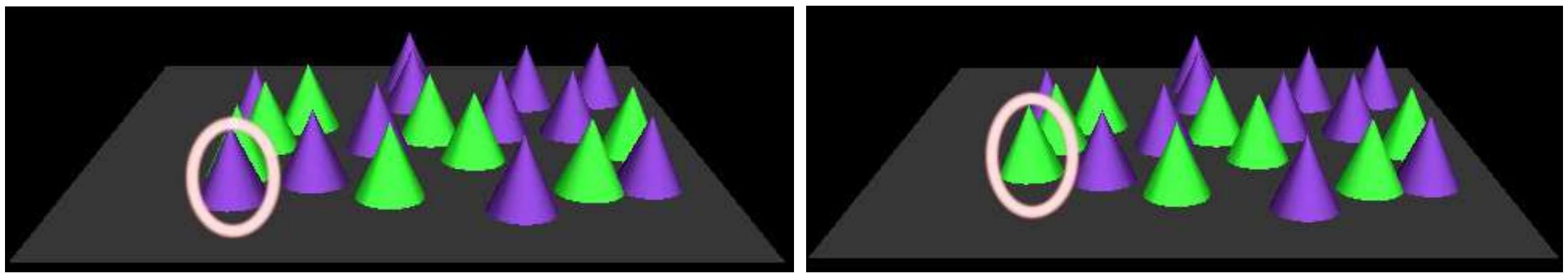

"the closest purple one on the far left side" "the green one right behind that one"

Figure 8: Example sequence of an anaphoric utterance

Participants employed spatial relations in 4% of the data (3% of the clean data).

### 2.3.6 Other

In addition to the phenomena listed in the preceding sections, participants used a small number of other description strategies. Some that occurred more than once but that we have not yet addressed in our computational model are selection by distance (lexicalised as "close to" or "next to"), selection by neighbourhood ("the green one surrounded by purple ones"), selection by symmetry ("the one opposite that one"), and selection by something akin to local connectivity ("the lone one"). There are also additional types of groupings, for example grouping by linearity ("the row of green ones", "the three purple on a diagonal") and picking out objects within a group by number ("the second one from the left in the row of five purple ones") that we do not cover here. Each of these strategies occurs less often in the data than anaphora does (it occurs in 4% of utterances, see the previous section).

We annotated 13% of our data as containing descriptive strategies other than the ones covered in the preceding sections. However, these other devices are often combined with the phenomena covered here. We marked 15% of our data as containing errors. Errors come in the form of repairs by the speaker, as faulty utterance segmentation by our speech segmenter, or through the misaligning of utterances with scenes by our system.

There are also a few instances of participants composing semantic phenomena in ways that we do not handle. There were two instances of combining spatial relations ("the one below and to the right") and a few instances of specifying groups by spatial extrema and regions ("the group of the purple ones on the left"). We did not count these as "other" in our evaluation, but rather we counted them as errors; the reported success rate is correspondingly lower.

### 2.4 Summary

The preceding sections catalogue strategies participants employed in describing objects. A computational system that understands utterances using these strategies must fulfill the following requirements:

- The system must have access to the visual scene and be able to compute visual features like those used by human speakers: natural groupings, inter-object distances, orderings and spatial relations
- It must have a robust language parsing mechanism that discovers grammatical patterns associated with descriptive strategies
- Feeding into the parsing mechanism must be a visually grounded lexicon; each entry in this lexicon must carry information as to which descriptive strategies it takes part in, and how these descriptive strategies combine with others
- The semantic interpretation and composition machinery must be embedded into the parsing process
- The system must be able to interpret the results of parsing an utterance and make a best guess as to which object the whole utterance describes

We go on to describe our system's understanding framework, consisting of the visual system, grounded lexical entries and the parser in Section 3. In Section 4 we discuss the modules we implemented to understand human descriptive strategies.

## 3. The Understanding Framework

In this section we describe the components of our Bishop understanding system in detail, with emphasis on how they fit together to work as a visually grounded understanding system. We cover in turn Bishop's vision system, its parser and lexicon and give a short overview of how our implementation of descriptive strategies fits into the framework.

### 3.1 Synthetic Vision

Instead of relying on the information we use to render the scenes in Bishop, which includes 3D object locations and the viewing angle, we implemented a simple synthetic vision algorithm. This algorithm produces a map attributing each pixel of the rendered image to one of the objects or the background. In addition, we use the full colour information for each pixel drawn in the rendered scene. Our goal is to loosely simulate the view of a camera pointed at a scene of real world objects, the situation our robots find themselves in. We have in the past successfully migrated models from synthetic vision (Roy, 2002) to computer vision (Roy et al., 2002) and plan on a similar route to deploy Bishop. Obviously, many of the hard problems of object detection as well as lighting and noise robustness do not need to be solved in the synthetic case, but we hope that the transfer back to a robot's camera will be made easier by working from a 2D image. We chose to work in a virtual world for this project so that we could freely change colour, number, size, shape and arrangement of objects to elicit interesting verbal behaviours in our participants, without running into the limitations of object detection algorithms or field of view problems.

Given an input image in which regions corresponding to objects have been segmented, the features produced by the vision system are:

**average RGB colour** the average of the red, green and blue components of all the pixels attributed to an object

**centre of mass** the average of the $x$ and $y$ pixel positions of an object

**distance** the euclidean distance between pairs of objects' centres of mass

**groups** the groups of objects in the scene as determined by finding all sets of objects that contain more than one object, and in which each object is less than a threshold distance from another object in the group (distances are measured between centres of mass)

**pairs** the same as groups, but filtered to produce only groups of two objects

**triplets** the same as groups, but filtered to produce only groups of three objects

**convex hull** the set of pixels forming the smallest convex region enclosing a set of objects

**attentional vector sum (AVS)** The AVS is a spatial relation measure between to objects. At extreme parameter settings it measures one of two angles, that formed by the centres, the other formed by the closest points of two objects. We use a parameter setting ($\lambda = 0.7$) in between these two extremes, which produces an intermediate angle depending on the object's shape. The resulting direction is measured relative to a set of reference angles, in our system the four Cartesian vectors $(0, 1), (0, -1), (1, 0), (-1, 0)$ (Regier & Carlson, 2001).

### 3.2 Knowledge Representation

Objects are represented as integer IDs in our system. For each ID or set of IDs the vision system can compute the visual features described in Section 3.1 based on the corresponding set of pixels in the image. The distinguishing ID together with the visual features represents the system's total knowledge of the objects present in the scene. The system can further instantiate new objects in the vision system from the convex hull of groups of other objects. The system also remembers the ID of the object removed last, and can ask the vision system to perform a feature computation on the visual scene as it was before the object was removed.

Groups of objects have their own integer IDs so they can be treated as objects themselves (all visual features are available for them). Their IDs are stored together with a list of their constituent objects' IDs, so that groups can be broken apart when necessary.

Finally, as visible in the lexicon file in Appendix B, each lexical item is stored with a set of associated parameters. These parameters specify grammatical type, compositional arity and reference behaviour (what the word sense can be taken as referring to on its own: a single object, a group of objects or no objects.) Furthermore, the lexical item is associated with a semantic composer (see Sections 3.3 and 4) which store their own sets of parameters, such as those specifying a Gaussian together with its applicable dimensions in the case of probabilistic composers.

### 3.3 Lexical Entries and Concepts

Conceptually, we treat lexical entries like classes in an object oriented programming language. When instantiated, they maintain an internal state that can be as simple as a tag identifying the dimension along which to perform an ordering, or as complex as multidimensional probability distributions. Each entry also has a function interface that specifies how it performs semantic composition. Currently, the interface definition consists of the number and arrangement of arguments the entry is willing to accept, whereas type mismatches are handled during composition rather than being enforced through the interface. Finally, each entry can contain a semantic composer that encapsulates the actual function to combine this entry with other constituents during a parse. These composers are described in-depth in Section 4. The lexicon used for Bishop contains many lexical entries attaching different semantic composers to the same word. For example, "left" can be either a spatial relation or an extremum. The grammatical structure detected by the parser (see the next Section) determines which compositions are attempted in a given utterance.

During composition, structures representing the objects that a constituent references are passed between lexical entries. We refer to these structures as *concepts*. Each entry accepts zero or more concepts, and produces zero or more concepts as the result of the composition operation. A concept lists the entities in the world that are possible referents of the constituent it is associated with, together with real numbers representing their ranking due to the last composition operation. A composer can also mark a concept as referring to a previous visual scene, to allow for anaphoric reference (see Section 4.5). It also contains flags specifying whether the referent should be a group of objects or a single object ("cones" vs. "cone"), and whether it should uniquely pick out a single object or is ambiguous in nature ("the" vs. "a"). These flags are used in the post-processing stage to determine possible ambiguities and conflicts.

Our lexicon, based on the development corpus, contains 93 words: 33 ADJ (adjectives), 2 CADJ (colour adjectives: "green", "purple"), 38 N (nouns), 2 REL (relative pronouns: "that", "which"), 1 VPRES (present tense verbs: "is"), 2 RELVPRES (relative pronoun/present tense verb combinations: "that's", "it's"), 1 ART ("the"), 3 SPEC (adjective specifiers: "right" (as in "right above"), "just"), 7 P (prepositions), 4 specific prepositions (POF, PAT, and two versions of PIN). The complete lexicon specification is reproduced in Appendix B.

### 3.4 Parsing

While in previous work we have used Markov models to parse and generate utterances (Roy, 2002), we here employ to context free grammars. These grammars naturally let us specify local compositional constraints and iterative structures. Specifically, they allow us to naturally perform grounded semantic composition whenever a grammar rule is syntactically complete, producing partial understanding fragments at every node of the parse tree. The parse structure of an utterance thus dictates which compositions are attempted. We use a bottom-up chart parser to guide the interpretation of phrases (Allen, 1995). Such a parser has the advantage that it employs a dynamic programming strategy to efficiently reuse already computed subtrees of the parse. Furthermore, it produces all sub-components of a parse and thus produces a useable result without the need to parse to a specific symbol. By using a dynamic programming approach we are assuming that meanings of parts can be assembled into meanings of wholes. We are not strictly committed to this assumption and in the future will consider backtracking strategies as necessary. Also note that due to the fact that our framework often produces functions to be applied at later stages of interpretation (see section 4) we avoid some possible overcommitting decisions (excluding the correct referent at an early stage of understanding).

Bishop performs only a partial parse, a parse that is not required to cover a whole utterance, but simply takes the longest referring parsed segments to be the best guess. Unknown words do not stop the parse process. Rather, all constituents that would otherwise end before the unknown word are taken to include the unknown word, in essence making unknown words invisible to the parser and the understanding process. In this way we recover essentially all grammatical chunks and relations that are important to understanding in our restricted task. For an overview of related partial parsing techniques, see the work of Abney (1997).

The grammar used for the partial chart parser is shown in Figure 1. Together with the grammar rules the table shows the argument structures associated with each rule. In the given grammar there is only one argument structure per rule, but there can be any number of argument structures. In the design of our grammar we were constrained by the compositions that must occur when a rule is applied. This can especially be seen for prepositional phrases, which must occur in a rule with the noun phrase they modify. The chart parser incrementally builds up rule fragments in a left to right fashion during a parse. When a rule is syntactically complete, it checks whether the composers of the constituents in the tail of the rule can accept the number of arguments specified in the rule (as shown in the last column of Table 1). If so, it calls the semantic composer associated with the constituent with the concepts yielded by its arguments to produce a concept for the head of the rule. If the compose operation fails for any reason (the constituent cannot accept the arguments or the compose operation does not yield a new concept) the rule does not succeed and does not produce a new constituent. If there are several argument structures (not the case in the final grammar shown here) or if a compose operation yields several alternative concepts, several instances of the head constituent are created, each with its own concept.

We provide an example chart produced by this grammar in Figure 10 in Section 5, as part of an example of the whole understanding process. The composition actions associated with each lexical item, and thus with each rule completion using this grammar, are listed in Appendix B.

### 3.5 Post-Parse Filtering

Once a parse has been performed, a post-parse filtering algorithm picks out the best interpretation of the utterance. First, this algorithm extracts the longest constituents from the chart that are marked as referring to objects, assuming that parsing more of the utter-ance implies better understanding. The filtering process then checks these candidates for consistency along the following criteria:

- All candidates must either refer to a group or to a single object
- If the candidates are marked as referring to an unambiguously specified single object, they must unambiguously pick a referent
- The referent in the specified single object case must be the same across all candidates
- If the candidates are marked as selecting a group, each must select the same group

| | | $T_0$ | $T_1$ | $T_2$ | $T_3$ | $T_4$ | $T_5$ | $T_6$ | Arg Structure |
|---|---|---|---|---|---|---|---|---|---|
| ADJ | ← | ADJ | ADJ | | | | | | $T_1(T_0)$ |
| NP | ← | ADJ | NP | | | | | | $T_0(T_1)$ |
| NP | ← | CADJ | N | | | | | | $T_0(T_1)$ |
| NP | ← | N | | | | | | | $T_0()$ |
| NP | ← | ART | NP | | | | | | $T_0(T_1)$ |
| NP | ← | NP | P | NP | | | | | $T_1(T_0, T_2)$ |
| NP | ← | NP | P | ART | N | POF | NP | | $T_3(T_0, T_5)$ |
| NP | ← | NP | RELVPRES | P | ART | N | POF | NP | $T_3(T_0, T_5)$ |
| NP | ← | NP | P | N | POF | NP | | | $T_2(T_0, T_4)$ |
| NP | ← | NP | REL | VPRES | NP | | | | $T_1(T_0, T_3)$ |
| NP | ← | NP | REL | P | NP | | | | $T_2(T_0, T_3)$ |
| NP | ← | NP | REL | VPRES | P | NP | | | $T_3(T_0, T_4)$ |
| NP | ← | NP | RELVPRES | P | NP | | | | $T_2(T_0, T_3)$ |
| NP | ← | NP | REL | VPRES | ADJ | | | | $T_3(T_0)$ |
| NP | ← | NP | RELVPRES | ADJ | | | | | $T_2(T_0)$ |
| NP | ← | NP | REL | CADJ | | | | | $T_2(T_0)$ |
| P | ← | SPEC | P | | | | | | $T_0(T_1)$ |
| P | ← | P | P | | | | | | $T_1()$ |
| P | ← | POF | | | | | | | $T_0()$ |

Table 1: Grammar used in BISHOP

If any of these consistency checks fail, the filtering algorithm can provide exact information as to what type of inconsistency occurred (within-group ambiguity, contradicting constituents, no object fulfilling the description), which constituents were involved in the inconsistency and which objects (if any) are referenced by each candidate constituent. In the future, we plan to use this information to resolve inconsistencies through active dialogue.

Currently, we enforce a best single object choice after the post processing stage. If the filtering yields a single object, nothing needs to be done. If the filtering yields a group of objects, we choose the best matching object (note that in this case we ignore the fact of whether the resulting concept is marked as referring to a group or a single object). If several inconsistent groups of referents remain after filtering, we randomly pick one object from the groups.

## 4. Semantic Composition

In this section we revisit the list of descriptive strategies from Section 2.3 and explain how we computationally capture each strategy and its composition with other parts of the utterance. Most of the composers presented follow the same composition schema: they take one or more concepts as arguments and yield another concept that references a possibly different set of objects. Concepts reference objects with real numbered values indicating the strength of the reference. Composers may introduce new objects, even ones that do not exist in the scene as such, and they may introduce new types of objects (e.g. groups of objects referenced as if they were one object). To perform compositions, each concept provides functionality to produce a single referent, or a group of referents. The single object produced is simply the one having maximum reference strength, whereas a group is produced by using a reference strength threshold below which objects are not considered as possible referents of this concept. The threshold is relative to the minimum and maximum reference strength in the concept. Most composers first convert an incoming concept to the objects it references, and subsequently perform computations on these objects.

Furthermore, composers can mark concepts as 'not referring', as 'referring to a single object' or as 'referring to a group of objects'. This is independent of the actual number of objects yielded by the concept, and can be used to identify misinterpretations and ambiguities. We currently use these flags to delay composition with arguments that do not refer to objects. For example, this constraint prevents "the left green" to cause any composition when green is considered to be an adjective. For such cases, new 'chaining' semantic composers are created that delay the application of a whole chain of compositions until a referring word is encountered. These chaining composers internally maintain a queue of composers. If during the argument for a composition operation does not refer to an object, both the composer producing the argument and the composer accepting it are pushed onto the queue. When the first referring argument is encountered, the whole queue of composers is executed starting with the new argument and proceeding backwards in the order the composers were encountered.

We plan to exploit these features further in a more co-operative setting than the one described here, where the system can engage in clarifying dialogue with the user. We explain in Section 3.5 how we converged on a single object reference in the task under discussion here, and what the other alternatives would be.

### 4.1 Colour - Probabilistic Attribute Composers

As mentioned in Section 3.1, we chose not to exploit the information used to render the scene, and therefore must recover colour information from the final rendered image. This is not a hard problem because Bishop only presents virtual objects in two colours. The renderer does produce colour variations in objects due to different angles and distances from light sources and the camera. The colour average for the 2D projection of each object also varies due to occlusion by other objects. In the interest of making our framework more easily transferable to a noisier vision system, we worked within a probabilistic framework. We separately collected a set of labelled instances of "green" and "purple" cones, and estimated a three dimensional Gaussian distribution from the average red, green and blue values of each pixel belonging to the example cones.

When asked to compose with a given concept, this type of probabilistic attribute composer assigns each object referenced by the source concept the probability density function evaluated at the measured average colour of the object.

### 4.2 Spatial Extrema and Spatial Regions - Ordering Composers

To determine spatial regions and extrema, an ordering composer orders objects along a specified feature dimension (e.g. x coordinate relative to a group) and picks referents at an extreme end of the ordering. To do so, it assigns an exponential weight function to objects according to

$$\gamma^{i(1+v)}$$

for picking minimal objects, where $i$ is the object's position in the sequence, $v$ is its value along the feature dimension specified, normalized to range between 0 and 1 for the objects under consideration. The maximal case is weighted similarly, but using the reverse ordering subtracting the fraction in the exponent from 2,

$$\gamma^{(i_{\max}-i)(2-v)}$$

where $i_{\max}$ is the number of objects being considered. For our reported results $\gamma = 0.38$. This formula lets referent weights fall off exponentially both with their position in the ordering and their distance from the extreme object. In that way extreme objects are isolated except for cases in which many referents cluster around an extremum, making picking out a single referent difficult. We attach this type of composer to words like "leftmost" and "top".

The ordering composer can also order objects according to their absolute position, corresponding more closely to spatial regions rather than spatial extrema relative to a group. The reference strength formula for this version is

$$\gamma^{(1+\frac{d}{d_{\max}})}$$

where $d$ is the euclidean distance from a reference point, and $d_{\max}$ the maximum such distance amongst the objects under consideration.

This version of the composer is attached to words like "middle". It has the effect that reference weights are relative to absolute position on the screen. An object close to the centre of the board achieves a greater reference weight for the word "middle", independently of the position of other objects of its kind. Ordering composers work across any number of dimensions by simply ordering objects by their Euclidean distance, using the same exponential falloff function as in the other cases. The ordering composer for "middle", for example, computes distance from the board centre to the centres of mass of objects, and thus prefers those that are centred on the screen.

### 4.3 Grouping Composers

For non-numbered grouping (e.g., when the describer says "group" or "cones"), the grouping composer searches the scene for groups of objects that are all within a maximum distance threshold from another group member. This threshold is currently set by hand based on a small number of random scenes in which the designers identified isolated groups and adjusted the threshold to correctly find them but not others. It only considers objects that are referenced by the concept it is passed as an argument. For numbered groups (“two”, “three”), the composer applies the additional constraint that the groups have to contain the correct number of objects. Reference strengths for the concept are determined by the average distance of objects within the group. We acknowledge that this approach to grouping is simplistic and we are currently investigating more powerful visual grouping algorithms that take topological features into consideration. In spite of our simple approach, we can demonstrate some instances of successfully understanding references to groups in Bishop.

The output of a grouping composer may be thought of as a group of groups. To understand the motivation for this, consider the utterance, “the one to the left of the group of purple ones”. In this expression, the phrase “group of purple ones” will activate a grouping composer that will find clusters of purple cones. For each cluster, the composer computes the convex hull (the minimal “elastic band” that encompasses all the objects) and creates a new composite object that has the convex hull as its shape. When further composition takes place to understand the entire utterance, each composite group serves as a potential landmark relative to “left”.

However, concepts can be marked so that their behaviour changes to split apart concepts refering to groups. For example, the composer attached to “of” sets this flag on concepts passing through it. Note that “of” is only involved in composition for grammar rules of the type NP ← NP P NP, but not for those performing spatial compositions for phrases like “to the left of”. Therefore, the phrase “the frontmost one of the three green ones” will pick the front object within the best group of three green objects.

### 4.4 Spatial Relations - Spatial Composers

The spatial semantic composer employs a version of the Attentional Vector Sum (AVS) suggested by Regier and Carlson (2001). The AVS is a measure of spatial relation meant to approximate human judgements corresponding to words like “above” and “to the left of” in 2D scenes of objects. It computes an interpolation between the angle between the centres of masses of the objects and the angle between the two closest points of the objects, in addition to a value depending on height relative to the top of the landmark object. Despite our participants talking about 2D projections of 3D scenes we found that the AVS distinguishes the spatial relations used in our data rather well when simply applied to the 2D projections. The participants often used spatial descriptors such as “below”, suggesting that they sometimes conceptualized the scenes as 2D. In a 3D setting we would expect to see consistent use of semantic patterns like “in front of” instead of “below”.

Given two concepts as arguments, the spatial semantic composer converts both into sets of objects, treating one set as providing possible landmarks, the other as providing possible targets. The composer then calculates the AVS for each possible combination of landmarks and targets. The reference vector used for the AVS is specified in the lexical entry containing the composer, e.g. $(0, 1)$ for “behind”. Finally, the spatial composer divides the result by the Euclidean distance between the objects’ centres of mass, to account for the fact that participants exclusively used nearby objects to select through spatial relations.

### 4.5 Anaphoric Composers

Triggered by words like "that" (as in "to the left of that one") or "previous", an anaphoric composer produces a concept that refers to a single object, namely the last object removed from the scene during the session. This object specially marks the concept as referring not to the current, but the previous visual scene, and any further calculations with this concept are performed in that visual context.

For example, when the parser calls upon the anaphoric composer attached to the lexical entry for "that" to provide an interpretation of "that one", this composer marks the produced concept as referring back to the previous visual scene, and sets the previously selected object as the only possible referent. Now consider another composer, say the spatial composer attached to "left" in "the one to the left of that one". When it asks for spatial relation features between the referents of "the one" and "that one", these spatial relation features (see Section 4.4) are computed on the previous visual scene with the object that was removed due to the previous utterance as the only possible landmark of the spatial relation.

## 5. Example: Understanding a Description

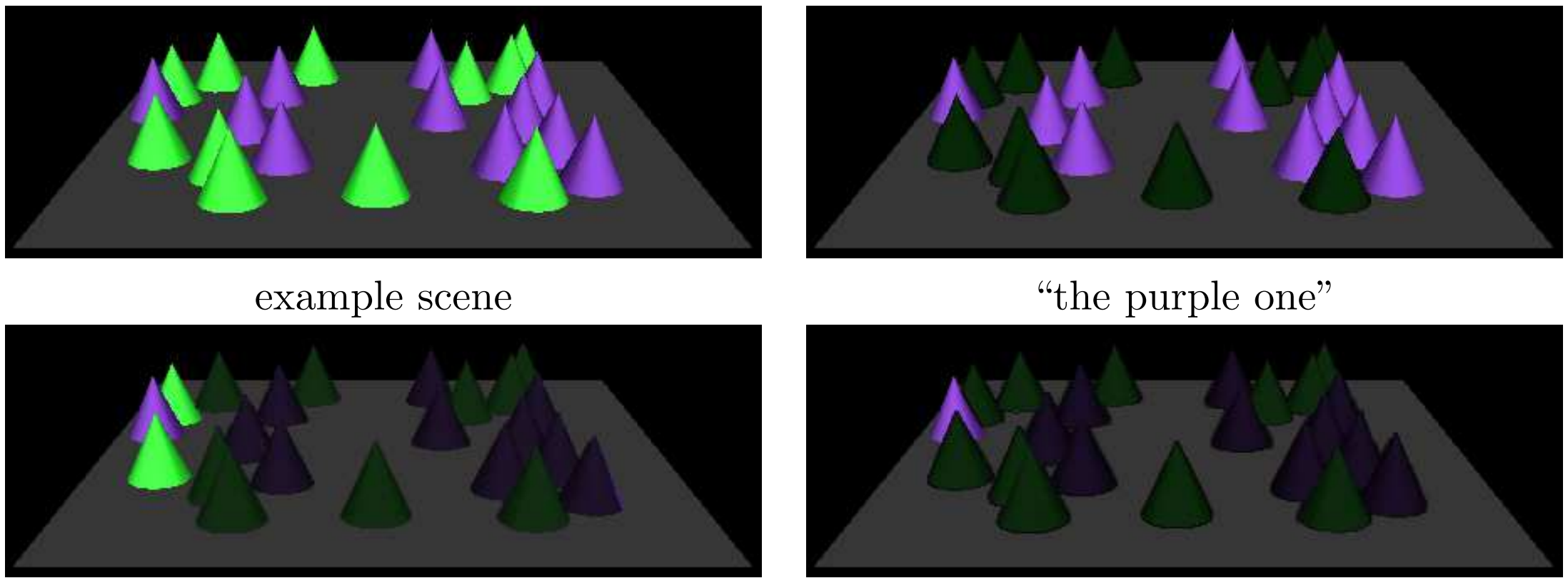

example scene — "the purple one"

"one on the left" — "the purple one on the left"

Figure 9: Example: "the purple one on the left"

To illustrate the operation of the overall system, in this section we step through some examples of how Bishop works in detail. Consider the scene in the top left of Figure 9, and the output of the chart parser for the utterance, "the purple one on the left" in Figure 10. Starting at the top left of the parse output, the parser finds "the" in the lexicon as an ART (article) with a selecting composer that takes one argument. It finds two lexical entries for "purple", one marked as a CADJ (colour adjective), and one as an N (noun). Each of them have the same composer, a probabilistic attribute composer marked as P(), but the adjective expects one argument whereas the noun expects none. Given that the noun expects no arguments and that the grammar contains a rule of the form NP← N, an NP (noun phrase) is instantiated and the probabilistic composer is applied to the default set of objects yielded by N, which consists of all objects visible. This composer call is marked

P(N) in the chart. After composition, the NP contains a subset of only the purple objects (Figure 9, top right). At this point the parser applies NP ← ART NP, which produces the NP spanning the first two words and again contains only the purple objects, but is marked as unambiguously referring to an object. S(NP) marks the application of this selecting composer called S.

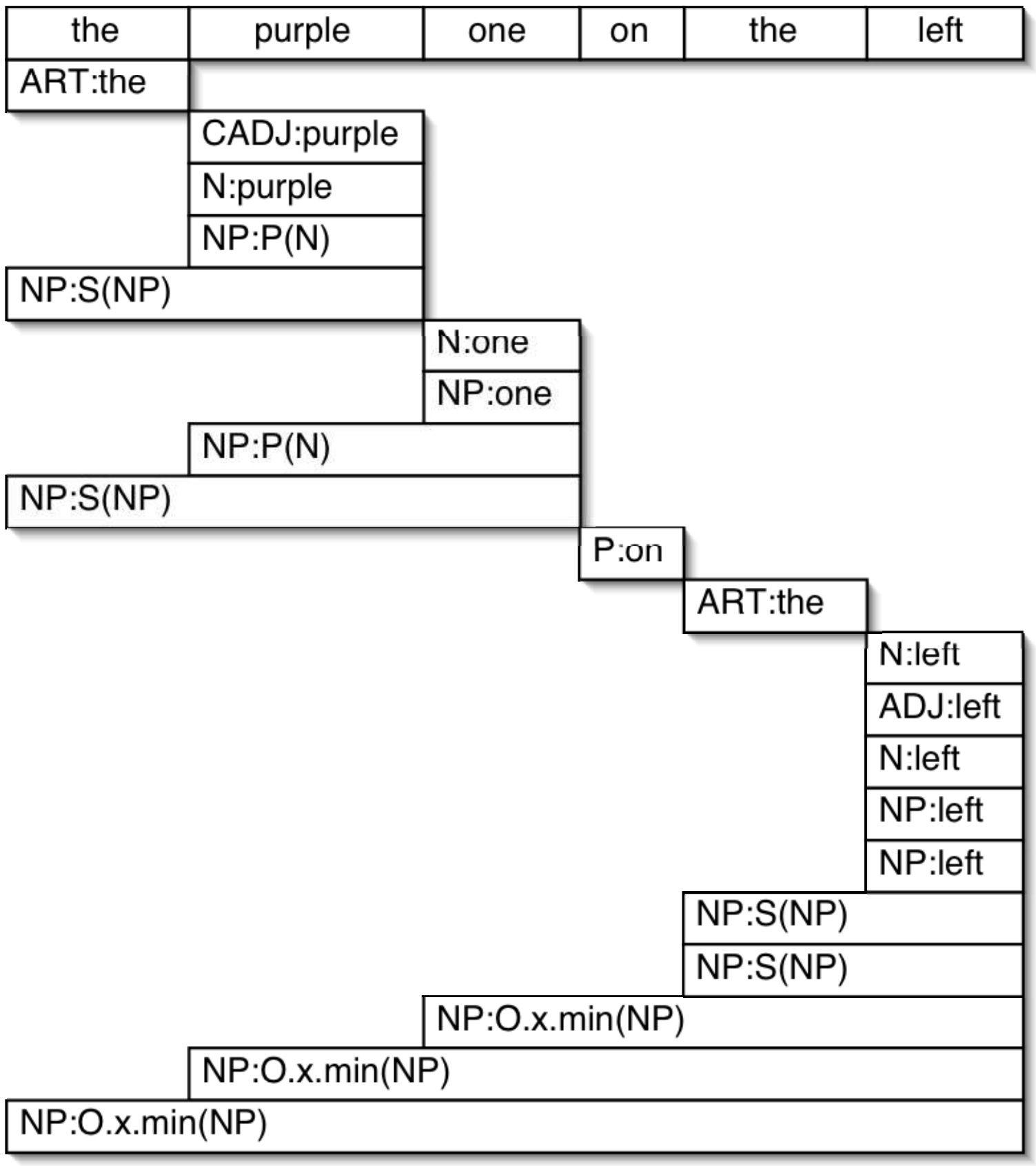

Figure 10: Sample parse of a referring noun phrase

The parser goes on to produce a similar NP covering the first three words by combining the "purple" CADJ with "one" and the result with "the". The "on" P (preposition) is left dangling for the moment as it needs a constituent that follows it. It contains a modifying semantic composer that simply bridges the P, applying the first argument to the second. After another "the", "left" has several lexical entries: in its ADJ and one of its N forms it contains an ordering semantic composer that takes a single argument, whereas its second N form contains a spatial semantic composer that takes two arguments to determine a target and a landmark object. At this point the parser can combine "the" and "left" into two possible NPs, one containing the ordering and the other the spatial composer. The first of these NPs in turn fulfills the need of the "on" P for a second argument according to NP ← NP P NP, performing its ordering compose first on "one" (for "one on the left"), selecting all the objects on the left (Figure 9, bottom left). The application of the ordering composer is denoted as $O.x.min$(NP) in the chart, indicating that this is an ordering composer ordering along the $x$ axis and selecting the minimum along this axis. On combining with "purple one", the same composer selects all the purple objects on the left (Figure 9, bottom right). Finally on "the purple one", it produces the same set of objects as "purple one", but marks the concept as unambiguously picking out a single object. Note that the parser attempts to use the second interpretation of "left" (the one containing a spatial composer) but fails because this composer expects two arguments that are not provided by the grammatical structure of the sentence.

## 6. Results and Discussion

In this section we first discuss our system's overall performance on the collected data, followed by a detailed discussion of the performance of our implemented descriptive strategies.

### 6.1 Overall Performance

For evaluation purposes, we hand-annotated the data, marking which descriptive strategies occurred in each example. Most examples use several reference strategies. In Table 2 we present overall accuracy results, indicating for which percentage of different groups of examples our system picked the same referent as the person describing the object. The first line in the table shows performance relative to the total set of utterances collected. The second one shows the percentage of utterances our system understood correctly excluding those marked as using a descriptive strategy that was not listed in Section 4, and thus not expected to be understood by Bishop. Some such examples are given in Section 2.3.6. The final line in Table 2 shows the percentage of utterances for which our system picked the correct referent relative to the clean development and testing sets, leaving out utterances marked as 'other' as well as those marked as containing some kind of error. As we defined earlier, this could be a speech error that was still understood by the human listener, or due to an error by the algorithm that pairs utterances with selection events. Additionally, relying on automatic speech segmentation sometimes merged utterances into one that should have been separate utterances. This mistakenly attributes the combination of two descriptions to one object selection and leaves another object selection without a corresponding utterance. Note, however, that due to our loose parsing strategy and the frequent redundancies in speaker's utterances our system was able to handle a good number of utterances marked as either 'other' or 'error'.

| Utterance Set | Accuracy - Development | Accuracy - Testing |
|---|---|---|
| All | 76.5% | 58.7% |
| All except 'Other' | 83.2% | 68.8% |
| All except 'Other' and 'Errors' (clean) | 86.7% | 72.5% |

Table 2: Overall Results

Using unconstrained speech primarily made writing a covering yet precise grammar difficult. This difficulty together with the loose parsing strategy made our system occasionally attempt compositions that are not supported by the grammatical structure of the utterance.

This overeager parsing strategy also produces a number of correct guesses that would not be found by a tighter grammar, and we found during development that the tradeoff often favoured looser parsing in terms of number of correct responses produced. Constructing the grammar is an obvious area to be addressed with a machine learning approach in the future. Using a speech segmenter together with an utterance reassembler produced very few errors because we used the successful selection event as a strong guideline for deciding which speech segments were part of a description. Errors of this type occur in less that 1% of the data.

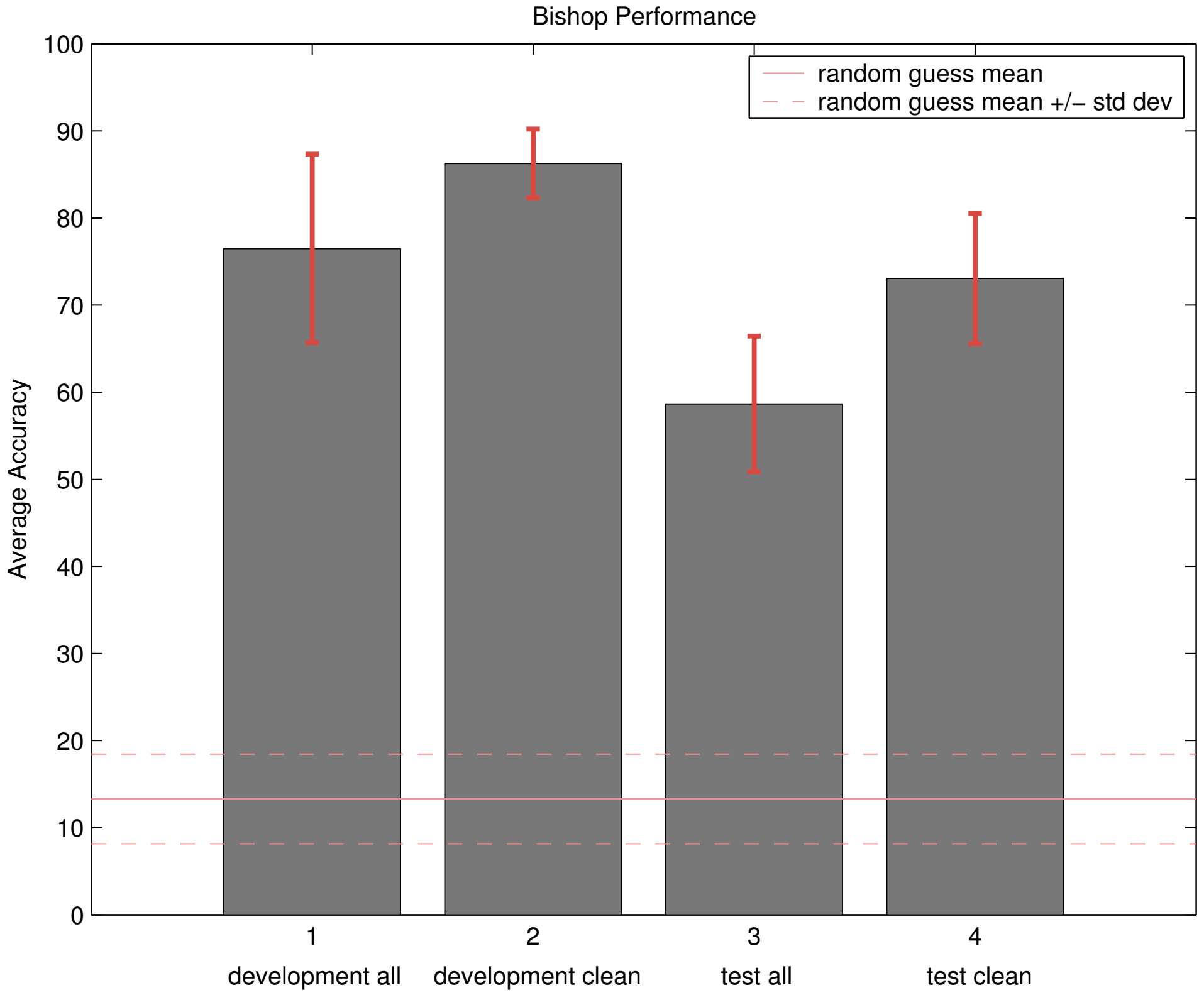

Figure 11: Results for the development and test corpora

Figure 11 graphs the results for each corpus and for a simulation of a uniform random selection strategy. Each bar shows the mean performance on a data set, with error bars delimiting one standard deviation. The figure shows results from left to right for the complete development corpus, the clean development corpus, the complete test corpus and the clean test corpus. Our system understands the vast majority of targeted utterances and performs significantly better than the random baseline. Given the unconstrained nature of the input and the complexity of the descriptive strategies described in Section 2.3 we consider this an important achievement.

Table 3 provides more detail for the various descriptive strategies and lists the percentage of correctly identified referents for utterances employing spatial extrema and regions, combinations of more than one spatial extremum, grouping constructs, spatial relations and anaphora. Note again that these categories are not mutually exclusive. We do not list separate results for the utterances employing colour terms because colour terms are not a source of errors (due to the synthetic nature of the vision system).

| Utterance Set | Accuracy - Development | Accuracy - Test |
|---|---|---|
| Spatial Extrema | 86.8% (132/152) | 77.4% (72/93) |
| Combined Spatial Extrema | 87.5% (49/56) | 75.0% (27/36) |
| Grouping | 34.8% (8/23) | 38.5% (5/13) |
| Spatial Relations | 64.3% (9/14) | 40.0% (8/20) |
| Anaphora | 100% (6/6) | 75.0% (3/4) |

Table 3: Detailed Results

Not surprisingly, Bishop makes more mistakes when errors are present or strategies other than those we implemented occur. However, Bishop achieves good coverage even in those cases. This is often a result of overspecification on the part of the describer. This tendency towards redundancy shows even in simple cases, for example in the use of "purple" even though only purple cones are left in the scene. It translates furthermore into specifications relative to groups and other objects when a simple 'leftmost' would suffice. Overspecification in human referring expressions is a well-known phenomenon often attributed to the incremental nature of speech production. Speakers may be listing visually salient characteristics such as colour before determining whether colour is a distinguishing feature of the intended referent (Pechmann, 1989).

The worst performance, that of the grouping composers, can be attributed both to the fact that the visual grouping strategy is too simplistic for the task at hand, and that this phenomenon is often combined in rather complex ways with other strategies. These combinations also account for a number of mistakes amongst the other composer that perform much better when combined with strategies other than grouping. We cover the shortcomings of the grouping composers in more detail in Section 6.2.3.

Mistakes amongst the descriptive strategies we cover have several causes:

**Overcommitment/undercommitment** Some errors are due to the fact that the interpretation is implemented as a filtering process without backtracking. Each semantic composer must produce a set of objects with attached reference strengths, and the next composer works from this set of objects in a strictly feedforward manner. During composition this strategy fails when the target object is left out at one stage (e.g. in "the leftmost one in the front", "leftmost" selects the leftmost objects, not including the obvious example of "front" that is not a good example of "leftmost"). It also fails when too many target objects are included (e.g. a poor example of "leftmost" is included in the set that turns out to be an ideal example of "front"). Estimating the group membership thresholds from the data will certainly decrease occurrence of these errors, but the real solution lies in a backtracking strategy combined with composers that are sensitive to the visual scenery beyond their immediate function. Such sensitive composers might take into account facts about the isolated nature of certain candidates as well as the global distribution of cones across the board. We discuss specific cases in which global and local visual context influence the interpretations of words in Section 6.2.

**Insufficient grammar** For some cases that contain many prepositional phrases (e.g. "the leftmost one in the group of purple ones on the right and to the bottom") our grammar was not specific enough to produce unambiguous answers. The grammar might attach "on the right" to the object rather than to the group of objects, not taking into account the biases in parsing that human listeners showed.

**Flawed composers** Some of the composers we have implemented are not sufficient to understand all facets of the corresponding human descriptive strategies. We will mention these problems in the following section.

### 6.2 Performance of Composers

We will now go reconsider each descriptive strategy and discuss the successes and failures of our composers that were designed to deal with each.

#### 6.2.1 Colour

Due to the simple nature of colour naming in the Bishop task, the probabilistic composers responsible for selecting objects based on colour made no errors.

#### 6.2.2 Spatial Regions and Extrema

Our ordering composers correctly identify 100% of the cases in which a participant uses only colour and a single spatial extremum in his or her description. We conclude that participants follow a process that yields the same result as ordering objects along a spatial dimension and picking the extreme candidate. Participants also favour this descriptive strategy, using it with colour alone in 38% of the clean data. Figure 3 provides examples of this type that our system handles without problems.

Description by spatial region occurs alone in only 5% of the clean data, and together with other strategies in 15% of the clean data. Almost all the examples of this strategy occurring alone use words like "middle" or "centre". The left image in Figure 12 exemplifies the use of "middle" that our ordering semantic composer models. The object referred to is the one closest to the centre of the board. We do not model the fact that human speakers only use this version of the descriptive strategy if there is an obvious single candidate object. The right image in Figure 12 shows a different interpretation of middle: the object in the middle of a group of objects. Note that the group of objects is linguistically not mentioned. Also note that within the group there are two candidate centre objects, and that the one in the front is preferred. Our composer only picks the correct object for this use of "middle" if the target object also happens to be the one closest to the centre of the board.

Figure 13 shows another use of the word "middle". This strategy seems related to the last one (picking the object in the middle of a group), however here the scene happens to be divided into two groups of objects with a single object in between them. Even though the object is in the back and not the closest one to the centre of the board, due to the visual

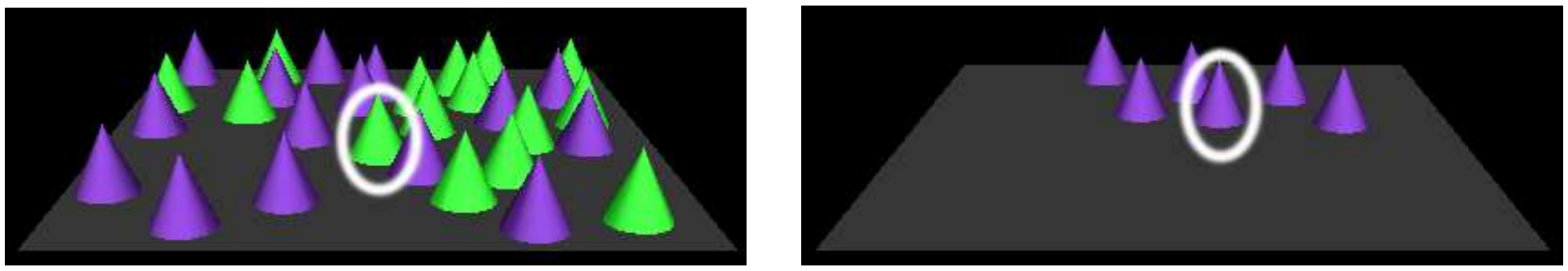

"the green one in the middle" "the purple cone in the middle"

Figure 12: Types of "middles" 1

context participants understand it to be the object in the "middle". Our composer fails in this case.

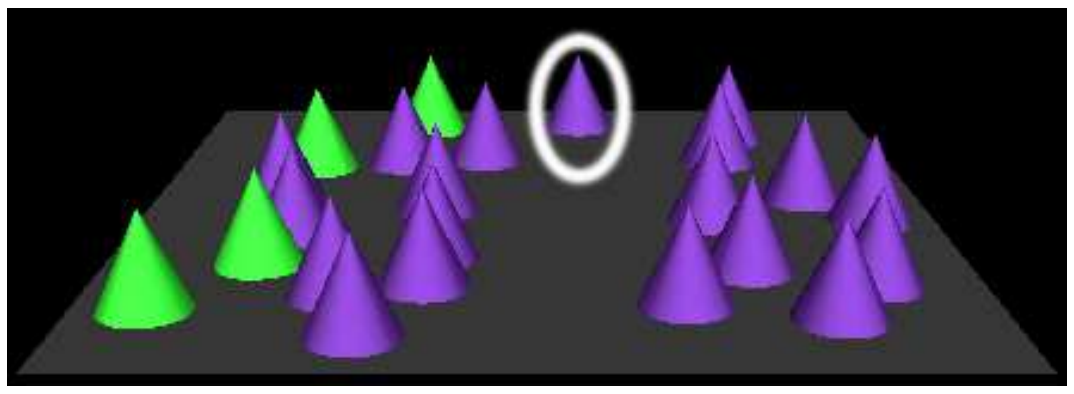

"the purple one in the middle"

Figure 13: Types of "middles" 2

Figure 14 shows a sequence of two scenes that followed each other during a data collection session. The first scene and utterance are a clear example of an extremum combined with a region specification, and our ordering composers easily pick out the correct object. In the next scene, the listener identified the leftmost object to be the one "right in the middle", despite the scene's similarity to the right image in Figure 12, where the "middle" object was in the middle of the group. We suspect that the use of "middle" in the scene before biases the understanding of "middle" as being relative to the board in this case, providing an example where not only visual, but also historical context influence the meanings of words. (Note that "right" in the utterance "right in the middle" is interpreted to have SPEC grammatical type by Bishop, and does not have a spatial role. See the grammar in Table 1.)

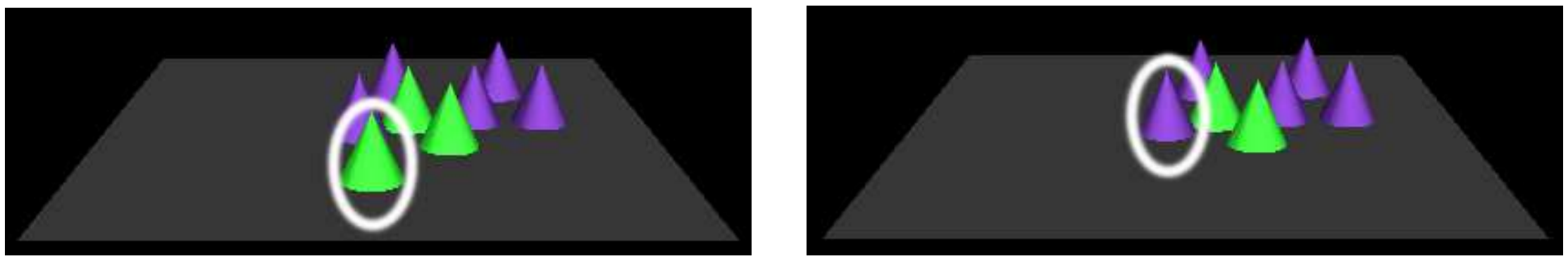

"the green one in the middle front" "the purple one right in the middle"

Figure 14: Types of "middles" 3

In summary, we can catalogue a number of different meanings for the word "middle" in our data that are linguistically indistinguishable, but depend on visual and historical context to be correctly understood. More generally, it is impossible to distinguish region-based uses from the various extrema-based uses of words based on the utterance alone in our data. We made the decision to treat "middle" to signify regions and "left", "top", etc. to signify extrema, but our examples of "middle" show that the selection of which meaning of these words to use depends on far subtler criteria such as global and local visual context, existence of an unambiguous candidate and past use of descriptive strategies.

Participants composed one or more spatial region or extrema references in 30% of the clean data. Our ordering composers correctly interpret 85% of these cases, for example those in Figure 4 in Section 2.3.2. The mistakes our composers make are usually due to overcommitment and faulty ordering. Figure 15 shows an example that could be interpreted as either problem (we indicate both the correct example and the object our system selects). We should note that this example comes from a non-native English speaker who often used "to" where native speakers would use "in". Our system selects the purple object closest to the back of the board instead of the indicated correct solution. This could be interpreted as overcommitment, because the composer for "back" does not include the target object, leaving the composer for "left" the wrong set of objects to choose from. A better explanation perhaps is that the ordering of the composers should be reversed in this case, so that the composer for "back" should take the objects selected by "left" as input. However, this ordering violates the far more common left-to-right ordering of region and extrema strategies in our data, which we selected to implement in our system. The question thus becomes what causes the difference in ordering in cases like the one in Figure 15. Once again, we suspect that visual context plays a role. Perhaps it is clear to the listener here that the double spatial specification would be an overspecification for the object our system selects (it is simply "the purple one in the back"). In response, the listener may seek an object that needs the full utterance, such as the true target. However, this analysis is hard to combine with the very common trend towards overspecification on the part of the speaker, leaving us with the need to run a more focused study of these phenomena to pin down the factors that play a role in their interpretation.

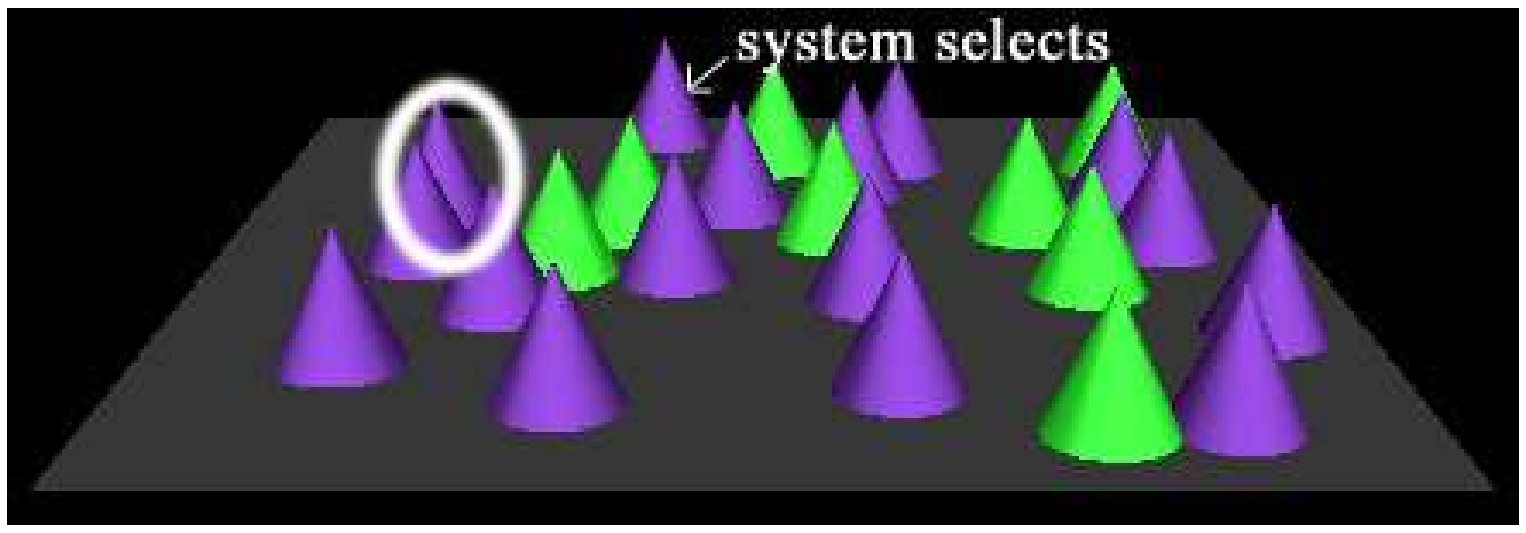

"purple cone to the back on the left side"

Figure 15: Misinterpreted utterance using composed extrema

#### 6.2.3 Grouping

Our composers implementing the grouping strategies used by participants are the most simplistic of all composers we implemented, compared to the depth of the actual phenomenon of visual grouping. The left scene in Figure 16 shows an example our grouping composer handles without a problem. The group of two cones is isolated from all other cones in the example, and thus is easily found by our distance thresholding algorithm. In contrast, the right scene depicts an example that would require much greater sophistication to find the correct group. The target group of three cones is not visually isolated in this scene, requiring further criteria like colinearity to even make it a candidate. Furthermore, there is a second colinear group of three cones that would easily be the best example of a "row of three purple cones" in the absence of the target group. It is only the target group's alignment with the vertical axis that let it stand out more as a "row" and make it the most likely interpretation. Our algorithm currently fails to include such grouping hints, and thus fails to pick the correct answer in this scene. Note that such hints are not always linguistically marked as they are here (through "row"), but often colinearity is silently assumed as holding for groups, making our simple grouping operator fail. A rich source of models for possible human grouping strategies like co-linearity comes from research in Gestalt Grouping (Wertheimer, 1999).

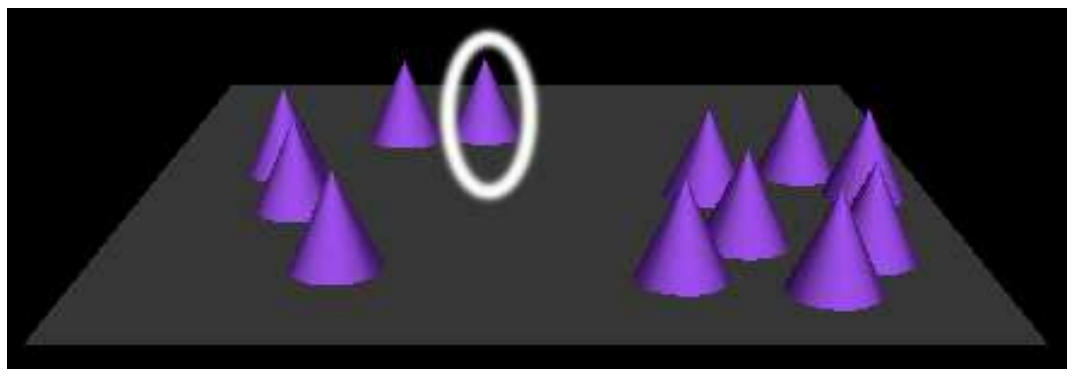

"the cone on the right in the pair of cones"

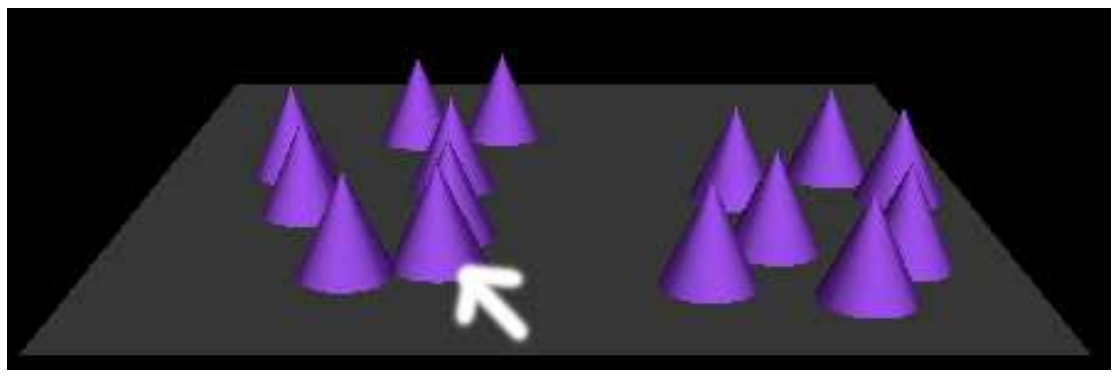

"the purple cone on at the front of the row of three purple cones"

Figure 16: Easy and hard visual grouping

#### 6.2.4 Spatial Relations

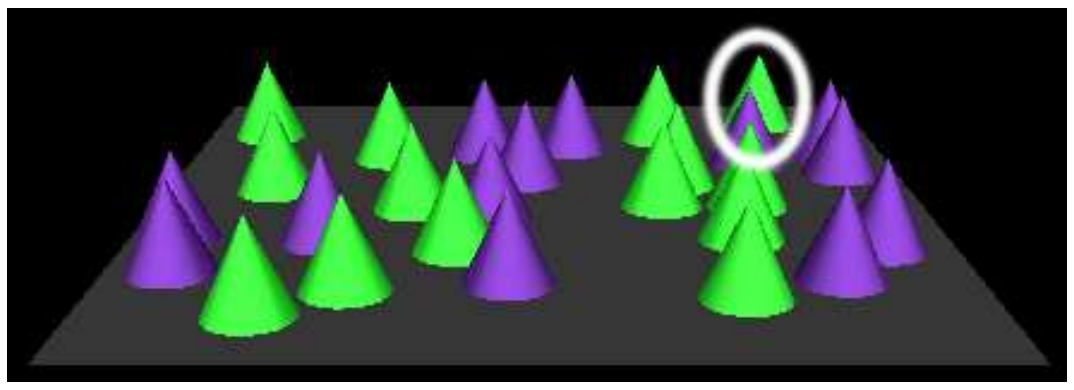

"the green cone behind the purple cone"

Figure 17: Successful spatial relation understanding

The AVS measure divided by distance between objects corresponds very well to human spatial relation judgements in this task. All the errors that occur in utterances that contain spatial relations are due to the possible landmarks or targets not being correctly identified (grouping or region composers might fail to provide the correct referents). Our spatial relation composer picks the correct referent in all those cases where landmarks and targets are the correct ones, for example in Figure 17. Also see the next section for a further correct example of spatial relations. Obviously, there are types of spatial relations such as relations based purely on distance and combined relations ("to the left and behind") that we decided not to cover in this implementation, but that occur in the data and should be covered in future efforts.

#### 6.2.5 Anaphora

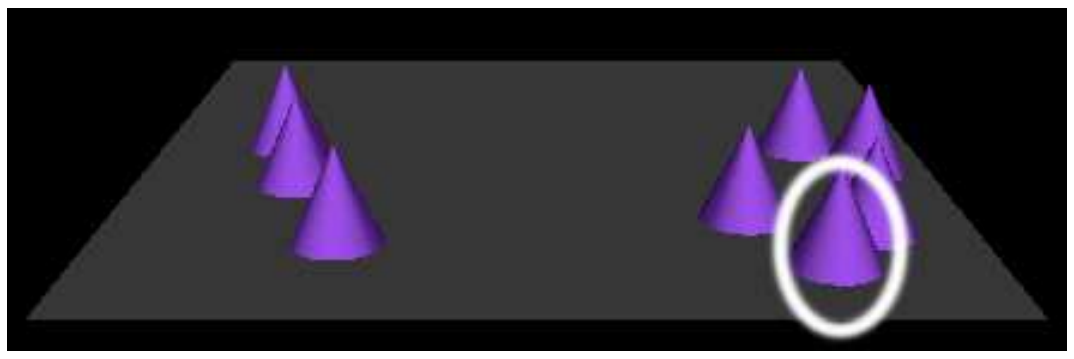

"the cone at the right front"

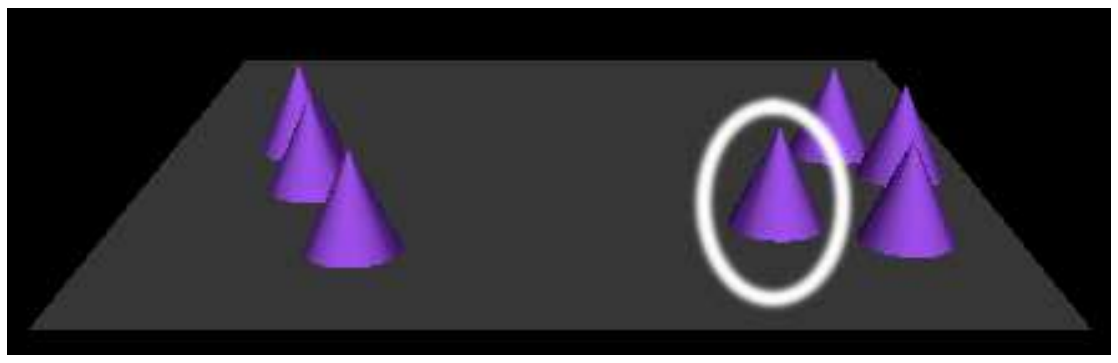

"the cone behind that on the left"

Figure 18: Successful Anaphora Understanding

Our solution to the use of anaphora in the Bishop task performs perfectly in replicating reference back to a single object in the clean data. This reference is usually combined with a spatial relation in our data, as in Figure 18. Due to the equally good performance of our spatial relation composer, we cover all cases of anaphora in the development data. However, there are more complex variants of anaphora that we do not currently cover, for example reference back to groups of objects such as in the sequence in Figure 19, which follows the right example in Figure 16.

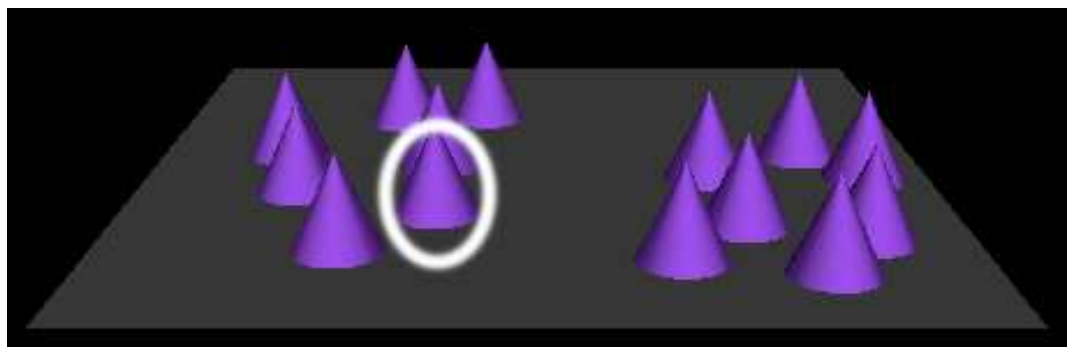

"the next cone in the row"

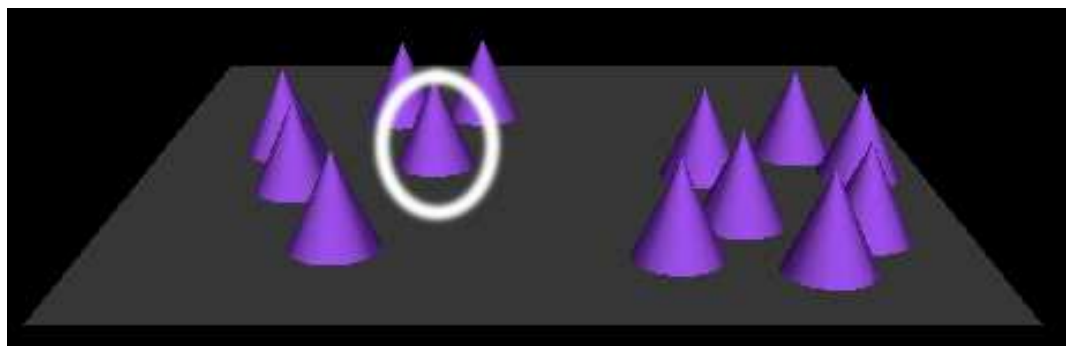

"the last cone in the row"

Figure 19: Group Based Anaphora

## 7. Future Directions

Given the analysis of Bishop's performance, there are several areas of future improvements that may be explored. The descriptive strategies we classified as 'other' should be understood by the computational system:

**Distance** A simple implementation to understand this strategy has the grammatical behaviour of our spatial relation composers, but uses an inverted distance measure to score target objects.

**Symmetry** Selection by symmetry only occurred for symmetry across the horizontal centre of the board in our data. We thus propose to mirror the landmark object across the horizontal centre, and scoring possible targets by their inverted distance to this mirror image.

**Numbered grouping** We limited ourselves to groups of two and three objects, but the same algorithm should work for higher numbers.

**In-group numbering** A descriptive strategy like "the second in the row" can be understood with a slight modification to our ordering composer that can put the peak of its exponential distribution not only at the ends and at the middle of the sequences, but rather at arbitrary points.

**Connectivity** A simple way to understand "the lonely cone" could measure distance to the closest objects within the group of possible referents. An better solution might construct a local connectivity graph and look for topologically isolated objects.

Furthermore, we have already mentioned several areas of possible improvement to the existing system due to faulty assumptions:

**Individual composers** Every one of our semantic composers attempts to solve a separate hard problem, some of which (e.g. grouping and spatial relations) have seen long lines of work dedicated to more sophisticated solutions than ours. The individual problems were not the emphasis of this paper. We believe that improvements in their implementation will improve our system as a whole, if not as much as the following possible techniques.

**Backtracking** The lack of backtracking in Bishop should be addressed. If a parse does not produce a single referent, backtracking would provide an opportunity to revise or "loosen" the decisions made at various stages of interpretation until a referent is produced.

**Visual context semantics** Backtracking only solves problems in which the system knows that it has either failed to obtain an answer, or knows that the answer it produced is an unlikely one. However, there are numerous examples in the data where one interpretation of an utterance produces a perfectly likely answer according to our measurements, for example "middle" finds an object at the exact centre of the screen. In many scenes this interpretation produces the correct answer, and a measurement relative to the other objects would produce a wrong one. However, we observe that participants only interpret "middle" in this way if there is no obvious structure to the rest of the scene. If by chance the scene is divided into a group of objects on the left and a group of objects on the right, "middle" will reliably refer to any isolated object in between these to groups, even if another object is closer to the actual centre of the screen. A future system should take into account local and global visual context during composition to account for these human selection strategies.

**Lexical entries** We have made the assumption that lexical entries are word-like entities that contain encapsulated semantic information. Even in our relatively constrained task, this is a somewhat faulty assumption. For example, the ambiguity in a word like "of" is only resolved by careful design of our grammar (see section 4.3), but it may be useful to treat entire phrases such as "to the left of" as single lexical entries, perhaps with its own grammar to replace "left" with other spatial markers (Jackendoff has proposed absorbing most rules of syntax into the lexicon, see Jackendoff, 2002).

**Dialogue** By constructing the parse charts we obtain a rich set of partial and full syntactic and semantic fragments offering explanations for parts of the utterance. At present, we largely ignore this information-rich resource when selecting the best referent. A more successful approach might entail backtracking and revision as described above, but also engage in clarification dialogue with the human speaker. The system could use the fragments it knows about to check the validity of its interpretation ("is this the group of green ones you mean?") or could simply disambiguate directly ("Which of these two do you mean?") followed by an explanation of its confusion in terms of the semantic fragments it formed.

Manual construction of a visually-grounded lexicon as presented here will be limited in accuracy due to various structural and parametric decisions that had to be manually approximated. Machine learning algorithms may be used to learn many of the parameter settings that were set by hand in this work, including on-line learning to adapt parameters during verbal interaction. Although some thresholds and probability distribution functions may be estimated from data using relatively straightforward methods, other learning problems are far more challenging. For example, learning new types of composers and appropriate corresponding grammatical constructs poses a difficult challenge for the future. Minimally, we plan to automate the creation of new versions of old composers (e.g. applied to different dimensions and attributes). Moving beyond this, it is not clear how, for example, the set handling functionality used to determine groups of referents can expand automatically and in useful ways. It will also be interesting to study how people learn to understand novel descriptive strategies.

We are also continuing work in applying our results from grounded language systems to multimodal interface design (Gorniak & Roy, 2003). We recently demonstrated an application of the Bishop system as described in this paper to the problem of referent resolution in the graphical user interface for a 3D modelling application, Blender (Blender Foundation , 2003). Using a Bishop/Blender hybrid, users can select sets of objects and correct wrong mouse selections with voice commands such as "select the door behind that one" or "show me all the windows".

## 8. Summary

We have presented a model of visually-grounded language understanding. At the heart of the model is a set of lexical items, each grounded in terms of visual features and grouping properties when applied to objects in a scene. A robust parsing algorithm finds chunks of syntactically coherent words from an input utterance. To determine the semantics of phrases, the parser activates semantic composers that combine words to determine their joint reference. The robust parser is able to process grammatically ill-formed transcripts of natural spoken utterances. In evaluations, the system selected correct objects in response to utterances for 76.5% of the development set data, and for 58.7% of the test set data. On clean data sets with various speech and processing errors held out, performance was higher yet. We suggested several avenues for improving performance of the system including better methods for spatial grouping, semantically guided backtracking during sentence processing, the use of machine learning to replace hand construction of models, and the use of interactive dialogue to resolve ambiguities. In the near future, we plan to transplant Bishop into an interactive conversational robot (Hsiao et al., 2003), vastly improving the robot's ability to comprehend spatial language in situated spoken dialogue.

## Acknowledgments

Thanks to Ripley, Newt and Jones.

## Appendix A. Utterances in the Test Data Set

The following are the 179 utterances we collected as our test data set. They are presented in the correct order and as seen by the understanding system. This means that they include errors due to faulty speech segmentation as well as due to the algorithm that stitches over-segmented utterances back together.

```
the green cone in the middle
the purple cone behind it
the purple cone all the way to the left
the purple cone in the corner on the right
the green cone in the front
the green cone in the back next to the purple cone
the purple cone in the middle front
the purple cone in the middle
the frontmost purple cone
the green cone in the corner
the most obstructed green cone
the purple cone hidden in the back
the purple cone on the right in the rear
the green cone in the front
the solitary green cone
the purple cone on at the front of the row of three purple cones
the next cone in the row
the last cone in the row
the cone on the right in the pair of cones
the other cone
the cone closest to the middle in the front
the cone in the right set of cones furthest to the left
the cone at the right front
```

```
the cone behind that on the left
the frontmost left cone
the backmost left cone
the solitary cone
the cone on in the middle on the right
the front cone the cone
the frontmost green cone
the green cone in the front to the right of the purple cone
the green cone at the back of the row of four
the cone the green cone behind the purple cone
the purple cone behind the row of three green cones
the frontmost green cone on the right
the green cone in the corner in the back
the green cone in the back
the purple cone in the back to the right
the green cone at the front left
purple cone behind it
the purple cone behind that one
the green cone behind the purple cone
the green cone in between two purple cone
the purple cone in the front
the purple cone touching the green cone
the green cone in the front
purple cone at the left
the green cone in the back left
the purple cone in the middle in front of the other two purple cones
the purple cone to the left of the four green cones
the purple cone to the left of that leftmost
green cone
frontmost green cone
rear cone
rightmost cone
rearmost cone
the left green cone
the purple cone the green cone
the green cone
the furthestmost green cone in the exact middle
the frontmost green cone
the rightmost green cone in the clump of four green cones to the right
the green cone in front of the two purple cones near the left
the green cone between the two purple cones in the back middle
the frontmost purple cone
the leftmost of the two purple cones on the right I mean on the left,
   sorry so that was the leftmost of the two purple cones on the left side
the green cone on the left halfway back
```

```
the frontmost green cone in front of a purple cone
the most middle purple cone
the green cone on the most left to the most left
the green cone in the middle in front of the other green cone
the only green cone on the left
the furthestmost purple cone on the left
the furthest green cone
the leftmost green cone
the leftmost purple cone
the middle green cone
the green cone between the two other green cones
the frontmost purple cone
the backmost purple cone
the green cone between the two purple cones nearest to the front
the leftmost purple cone
the green cone in the front
the green cone
the frontmost of the two back purple cones
the rightmost purple cone
the leftmost purple cone
the purple cone in the front
the last purple cone
the frontmost purple cone in the clump of five purple cones
   on the right
backmost green cone
the backmost purple cone
the green cone directly in front of the purple cone
the purple cone behind the green cone on the left
the green cone behind the purple cone on the left
the leftmost of the two left back corner green cones
the rightmost purple cone
the middle cone behind the frontmost purple cone
the green cone on the left front corner
the purple cone on the right back corner
the third green cone in the line of green cones near the middle
the green cone between the two purple cones near the back
the green cone in the back left
the only green cone in the back
the green cone behind the frontmost green cone
the frontmost green cone
the only green cone
the last of the line of four purple cones
the centre purple cone in the three cones on the left
the purple cone between the two purple cones
the middle purple cone
```

the leftmost purple cone
the middle purple cone
the front left purple cone
the front right purple cone
the second of the four purple cones
the middle purple cone
the purple cone on the left
the last purple cone
the green one in the middle all the way at the back
the purple one it's all the way in the middle but a little but
    to the left and all the way in the back
the green one in the middle in the front that's in front of another
    green one
the purple one in the middle that's behind the green one
on the on the right the purple one at the very front of the line of
    purple ones
on the left the green one in between two purple ones in a line
and on the left the purple one in the middle of a row of I mean in the middle
    of a line of three purple ones
the green one on the left that's hidden by a purple one
on the left the purple one that's all the way in the corner and it's separate
in the middle towards the right there's a line of purple ones and then
    there's a kink in the line and the one that's right where the lines turns
the purple one all the way on the right in the front
the purple one in the front in the middle
the green one in the middle
the purple in the front all the way on the right
and the rightmost green one
the leftmost green one
and then the last green one the last green one
the frontmost purple one on the right
the purple one in the back towards the left that's next to two other
    purple ones
the purple one in the back towards the right that's not part of a pair
the purple one in the front of a group on the right
the purple one in the middle that's in front of a group one the right
the purple one on the left all the way in the back
the purple one on the left that's behind another purple one
the purple one on the left that's in the front the purple one
on the left that's by itself
the purple one on the right that's hidden by two other purple ones
the purple one all the way in the back corner on the right
the purple one that's in front on the right and the last one
and the last one
the purple on in the front on the right

```
the only purple one all the way on the right
the green one on the right that's in the middle of a bunch
the green one all the way on the left that's almost totally obscured
the last purple one on the left that in a crooked line of purple ones
the first purple one on the left that's in a crooked line
the purple one all the way one the all the way in the back towards
      the left that's behind a green and a purple one all the way in
      the back
the purple one towards the back that's pretty much in the back but
      that's in front of a green and purple one
the purple one in the middle in the back
the purple one on the left that's furthest to the back
the green one in the middle that's furthest to the front
the purple one towards the in the middle but towards the left
      that's closest
in the middle the purple one that stands out that's closest
of the purple ones in the middle and towards the right the one
      at the corner
the purple one that's closest to the middle
all the way on the right the green one in the middle of a line of three
      green ones
and then all the way on the right the closest green one
all the way on the right the only close green one
the green one all the way in the right corner in the back
the purple one that's towards the back and the left corner
the purple one in the front left corner
the purple one near the middle that's not with another purple one
the purple one that's in front of another purple one
and the only purple one on the right
the only purple one on the left
the green one in the middle that's just behind another green one
and the closest green one in the middle the green one that's closest to
      the middle
the green one all the way in the back towards the right
the only close green one the only one left
the only one left
```

## Appendix B

The following specifies the complete lexicon used in Bishop in XML format. The initial comment explains the attributes of lexical entries.

`see online appendix file 'lexicon.xml'.`

## References

Abney, S. (1997). Part-of-speech tagging and partial parsing. In *Corpus-Based Methods in Language and Speech*, chap. 4, pp. 118–136. Kluwer Academic Press, Dordrecht.

Allen, J. (1995). *Natural Language Understanding*, chap. 3. The Benjamin/Cummings Publishing Company, Inc, Redwood City, CA, USA.

Bailey, D. (1997). *When push comes to shove: A computational model of the role of motor control in the acquisition of action verbs*. Ph.D. thesis, Computer science division, EECS Department, University of California at Berkeley.

Barker, C. (2002). The dynamics of vagueness. *Linguistics and Philosophy*, *25*, 1–36.

Blender Foundation (2003). Blender 3D graphics creation suite. http://www.blender3d.org.

Brown, M., Buntschuh, B., & Wilpon, J. (1992). SAM: A perceptive spoken language-understanding robot. *IEEE Transactions on Systems, Man and Cybernetics*, *6*(22), 1390–1402.

Brown-Schmidt, S., Campana, E., & Tanenhaus, M. K. (2002). Reference resolution in the wild. In *Proceedings of the Cognitive Science Society*.

Carletta, J., & Mellish, C. (1996). Risk-taking and recovery in task-oriented dialogue. *Journal of Pragmatics*, *26*, 71–107.

Desolneux, A., Moisan, L., & Morel, J. (2003). A grouping principle and four applications. *IEEE Transactions on Pattern Analysis and Machine Intelligence*, *255*(4), 508–513.

Dhande, S. (2003). A computational model to connect gestalt perception and natural language. Master's thesis, Massachusetts Institute of Technology.

Engbers, E., & Smeulders, A. (2003). Design considerations for generic grouping in vision. *IEEE Transactions on Pattern Analysis and Machine Intelligence*, *255*(4), 445–457.

Eugenio, B. D., Jordan, P. W., Thomason, R. H., & Moore, J. D. (2000). The agreement process: An empirical investigation of human-human computer-mediated collaborative dialogues. *International Journal of Human-Computer Studies*, *53*(6), 1017–1076.

Gorniak, P., & Roy, D. (2003). Augmenting user interfaces with adaptive speech commands. In *Proceedings of the International Conference for Multimodal Interfaces*.

Griffin, Z., & Bock, K. (2000). What the eyes say about speaking. *Psychological Science*, *11*, 274–279.

Haddock, N. (1989). Computational models of incremental semantic interpretation. *Language and Cognitive Processes*, *4*, 337–368.

Hsiao, K., Mavridis, N., & Roy, D. (2003). Coupling perception and simulation: Steps towards conversational robotics. In *Proceedings of the IEEE/RSJ International Conference on Intelligent Robots and Systems (IROS)*.

Jackendoff, R. (2002). What's in the lexicon?. In Noteboom, S., Weerman, F., & Wijnen (Eds.), *Storage and Computation in the Language Faculty*, chap. 2. Kluwer Academic Press.

Kyburg, A., & Morreau, M. (2000). Fitting words: vague words in context. *Linguistics and Philosophy*, *23*, 577–597.

Lammens, J. M. (1994). *A computational model of color perception and color naming*. Ph.D. thesis, State University of New York.

Landau, B., & Jackendoff, R. (1993). "what" and "where" in spatial language and spatial cognition. *Behavioural and Brain Sciences*, *2*(16), 217–238.

Miller, G., & Johnson-Laird, P. (1976). *Language and Perception*. Harvard University Press.

Nagao, K., & Rekimoto, J. (1995). Ubiquitous talker: Spoken language interaction with real world objects. In *Proceeding of the International Joint Conference on Artificial Intelligence*.

Narayanan, S. (1997). *KARMA: Knowledge-based Action Representations for Metaphor and Aspect*. Ph.D. thesis, Univesity of California, Berkeley.

Partee, B. H. (1995). Lexical semantics and compositionality. In Gleitman, L. R., & Liberman, M. (Eds.), *An Invitation to Cognitive Science: Language*, Vol. 1, chap. 11, pp. 311–360. MIT Press, Cambridge, MA.

Pechmann, T. (1989). Incremental speech production and referential overspecification. *Linguistics*, *27*, 89–110.

Pustejovsky, J. (1995). *The Generative Lexicon*. MIT Press, Cambridge, MA, USA.

Regier, T. (1996). *The Human Semantic Potential*. MIT Press.

Regier, T., & Carlson, L. (2001). Grounding spatial language in perception: An empirical and computational investigation. *Journal of Experimental Psychology: General*, *130*(2), 273–298.

Roy, D. (2002). Learning visually-grounded words and syntax for a scene description task. *Computer Speech and Language*, *16*(3).

Roy, D., Gorniak, P. J., Mukherjee, N., & Juster, J. (2002). A trainable spoken language understanding system. In *Proceedings of the International Conference of Spoken Language Processing*.

Roy, D., & Pentland, A. (2002). Learning words from sights and sounds: A computational model. *Cognitive Science*, *26*(1), 113–146.

Schuler, W. (2003). Using model-theoretic semantic interpretation to guide statistical parsing and word recognition in a spoken language interface. In *Proceedings of the Association for Computational Linguistics*.

Shi, J., & Malik, J. (2000). Normalized cuts and image segmentation. *IEEE Transactions on Pattern Analysis and Machine Intelligence*, *8*(22), 888–905.

Siskind, J. M. (2001). Grounding the lexical semantics of verbs in visual perception using force dynamics and event logic. *Journal of Artificial Intelligence Research*, *15*, 31–90.

Wertheimer, M. (1999). Laws of organization in perceptual forms. In *A source book of Gestalt psychology*, pp. 71–88. Routledge, New York.

Winograd, T. (1970). *Procedures as a representation for data in a computer program for understanding natural language*. Ph.D. thesis, Massachusetts Institute of Technology.

Yoshida, N. (2002). Utterance segmenation for spontaneous speech recognition. Master's thesis, Massachusetts Institute of Technology.